\documentclass{article}

\usepackage[preprint]{neurips_2026}

\usepackage{amsmath,amsfonts,bm}

\def\eqref#1{equation~\ref{#1}}

\def\1{\bm{1}}
\newcommand{\train}{\mathcal{D}}

\newcommand{\test}{\mathcal{D_{\mathrm{test}}}}

\DeclareMathAlphabet{\mathsfit}{\encodingdefault}{\sfdefault}{m}{sl}
\SetMathAlphabet{\mathsfit}{bold}{\encodingdefault}{\sfdefault}{bx}{n}

\newcommand{\KL}{D_{\mathrm{KL}}}

\DeclareMathOperator*{\argmin}{arg\,min}

\usepackage[utf8]{inputenc}
\usepackage[T1]{fontenc}
\usepackage{hyperref}
\usepackage{url}
\usepackage{booktabs}
\usepackage{amsfonts}
\usepackage{nicefrac}
\usepackage{microtype}
\usepackage{xcolor}

\usepackage{graphicx}
\usepackage{subcaption}
\usepackage{minitoc}

\usepackage{amsmath}
\usepackage{amssymb}
\usepackage{mathtools}
\usepackage{amsthm}

\usepackage[capitalize,noabbrev]{cleveref}

\theoremstyle{plain}
\newtheorem{theorem}{Theorem}[section]
\newtheorem{proposition}[theorem]{Proposition}
\newtheorem{lemma}[theorem]{Lemma}
\newtheorem{corollary}[theorem]{Corollary}
\theoremstyle{definition}
\newtheorem{definition}[theorem]{Definition}

\theoremstyle{remark}

\theoremstyle{example}
\newtheorem{example}[theorem]{Example}

\usepackage[textsize=tiny]{todonotes}

\usepackage{helvet}
\usepackage{courier}
\usepackage{tabularx}
\usepackage{mathabx}
\usepackage{varwidth}
\usepackage{bbm}
\usepackage{bm}
\usepackage{dsfont}
\usepackage{colortbl}
\usepackage[clock]{ifsym}
\usepackage{enumitem}
\usepackage{multicol}
\usepackage{algorithm}
\usepackage{algorithmic}
\usepackage{wrapfig}
\usepackage{float}
\usepackage{multirow}
\usepackage{setspace}
\usepackage{pifont}
\usepackage{array}
\usepackage{comment}
\usepackage{thm-restate}

\definecolor{sexybrown}{HTML}{8B4513}
\definecolor{pierCite}{HTML}{00008B}

\usepackage{tikz}
\usetikzlibrary{shapes,arrows,fit,positioning,backgrounds}
\usetikzlibrary{arrows.meta, positioning, decorations.markings}

\newcommand{\bV}{{\mathbf{V}}}
\newcommand{\bv}{{\mathbf{v}}}
\newcommand{\bZ}{{\mathbf{Z}}}

\newcommand{\bU}{{\mathbf{U}}}
\newcommand{\bu}{{\mathbf{u}}}
\newcommand{\bX}{{\mathbf{X}}}
\newcommand{\bx}{{\mathbf{x}}}
\newcommand{\bY}{{\mathbf{Y}}}
\newcommand{\by}{{\mathbf{y}}}

\newcommand{\bR}{{\mathbf{R}}}
\newcommand{\br}{{\mathbf{r}}}

\newcommand{\bS}{{\mathbf{S}}}

\newcommand{\Pa}{{\mathbf{Pa}}}

\newcommand{\pa}{{\mathbf{pa}}}

\newcommand{\cF}{\mathcal{F}}
\newcommand{\cG}{\mathcal{G}}
\newcommand{\cP}{\mathcal{P}}
\newcommand{\cM}{\mathcal{M}}

\newcommand{\cI}{\mathcal{I}}

\newcommand{\cS}{\mathcal{S}}
\newcommand{\cO}{\mathcal{O}}

\newcommand{\cX}{\mathcal{X}}

\newcommand{\cH}{\mathcal{H}}

\newcommand{\cL}{\mathcal{L}}
\newcommand{\cU}{\mathcal{U}}
\newcommand{\cY}{\mathcal{Y}}
\newcommand{\cV}{\mathcal{V}}
\newcommand{\cJ}{\mathcal{J}}
\newcommand{\cE}{\mathcal{E}}

\newcommand{\ERM}{\mathrm{ERM}}
\newcommand{\AD}{\mathrm{AD}}

\newcommand{\TR}{\mathrm{TR}}
\newcommand{\src}{\mathrm{src}}
\newcommand{\trg}{\mathrm{trg}}

\newcommand{\query}{\mathrm{Que}}
\newcommand{\key}{\mathrm{Key}}

\newcommand{\prob}{\Psi}

\newcommand{\ft}{\mathrm{ft}}
\newcommand{\adapt}{\mathrm{ft}}
\newcommand{\true}{\mathrm{true}}

\newcommand{\pTR}{\mathrm{pTR}}

\newcommand{\bbE}{\mathbb{E}}

\newcommand{\indep}{\perp \!\!\! \perp}

\usepackage{titletoc}
\newcommand\DoToC{%
  \setcounter{tocdepth}{2}
  \startcontents
  {\centering\Large\bfseries Supplementary Material\par}
  \vskip3pt\hrule\vskip5pt
  \printcontents{}{1}{}
  \vskip3pt\hrule\vskip5pt
}

\title{Adapting, Fast and Slow:\\ On Few-Shot Transportability of Compositions}

\author{%
  Kasra Jalaldoust { \normalfont and } Elias Bareinboim \\
  Causal Artificial Intelligence Lab\\
  Columbia University \\
  \texttt{\{kasra,eb\}@cs.columbia.edu}
}

\begin{document}

\maketitle

\begin{abstract}
Generalization across domains requires stable structure that links the source and target distributions. Building on causal transportability theory, we study a sequential prediction setting in which the target predictor can be represented as a circuit composed of causal mechanisms that are learnable from source data. We introduce two classes of transportability. \emph{Module transportability} captures the atomic case, where the target predictor is given by a mechanism learnable from a single source domain. \emph{Circuit transportability} generalizes this idea to target predictors obtained by composing several modules learned from source data, enabling zero-shot prediction even when no source mechanism directly predicts the target label. We study these classes of circuits under increasingly relaxed assumptions. First, we provide conditions under which the relevant circuits can be learned from source data alone, given causal knowledge about the source and target domains. We then relax these structural assumptions by allowing limited data from the target domain. In particular, we develop a supervised domain adaptation scheme that learns circuits without requiring explicit causal structure. The resulting few-shot guarantees tie the achievable error to the size of the smallest target circuit composable from modules learned from source data. Finally, we propose a gradient-based relaxation of the symbolic circuit search and evaluate it empirically, showing that it qualitatively tracks the predicted regimes of fast adaptation -- with and without process supervision over intermediate positions -- and slow adaptation when no source mechanism matches. A held-out arithmetic-composition experiment (Euclidean GCD composed from an arithmetic-primitive library) shows that the framework also delivers near-oracle predictive accuracy when the target's data-generating process is a real algorithm rather than a randomly drawn SCM.

\end{abstract}

\section{Introduction}

Machine learning deals with generalizing patterns from finite samples to the distribution that generates them. The classical sample-to-population guarantees \citep{Vapnik1991,vapnik1998statistical} rely on the assumption that the \emph{target domain}, where the solution will be evaluated, induces a data distribution identical to the \emph{source domain}, where training happens. In practice, performance can degrade sharply under even small qualitative differences between source and target. This problem is known broadly as \emph{distribution shift} in ML, and as generalizability or external validity in a broader scientific context. In particular, the \emph{domain generalization} task refers to a setting where the learner has access to (typically large) data collected from one or multiple source domains and no data from the target domain. \emph{Domain adaptation} is the related setting where the learner additionally has access to a small amount of labeled data from the target. Theoretical understanding of generalization across domains is challenging. Arbitrary differences between the source and target domains impose a barrier on learning, since the source data has no a priori bearing on the target task. A formal approach therefore requires positing a \emph{structure} that specifies how the target relates to the source. Given such structure, one can design algorithms that leverage only statistical associations from the source data which provably remain stable in the target, yielding prediction rules with out-of-distribution guarantees \citep{peters2016causal,rojas2018invariant,rothenhausler2021anchor}.

Domain adaptation in prediction tasks involving covariates $X$ and label $Y$ has been studied in the literature \citep{ben-david-2006,NIPS2007_42e77b63,conf/colt/MansourMR09,ben2010theory,Hanneke2010,Hanneke2019,hanneke2024more}, where various notions of divergence between the source and target $X$ distribution are used as proxies for domain-relatedness. Other work in this area leverages distributional assumptions relating source and target, e.g., \citet{blitzer2011, bendavid2003, baxter2000, baxter1997} where learning in the source yields smaller complexity for learning in the target, e.g., through learning a shared \emph{representation}. Adjacent lines on compositional generalization in seq2seq, algorithmic composition, and independent causal mechanisms are surveyed in \Cref{app:related-comp-gen}.

Humans are particularly effective at transferring knowledge across domains \citep{Lake2017,Marcus2001,Tenenbaum2011}, and causality is widely understood as a pillar of human understanding and decision making, especially under changing circumstances \citep{Gopnik2004,Schulz2004}. Principles of generalization to unseen domains from a causal perspective have been studied extensively under the rubric of \emph{transportability} \citep{pearl2011transportability,bareinboim2013transportability,correa2019statistical,correa2020gtr,Jalaldoust_Bareinboim_2024,jalaldoust2024partial}, and also through the lens of statistical invariances entailed by an implicit causal structure \citep{peters2016causal,magliacane2018domain,koyama2021invariance,li2018domain}. In DA, since \emph{some} target data is available, the learner can always discard the source data entirely and rely solely on the target. The theoretical question in DA is therefore not whether learning is possible, but how fast it can take place and how best to leverage the source data. In this paper we work in a \emph{sequential prediction setting}: the data at each domain consists of indexed positions $V_1, V_2, \dots$ over a discrete vocabulary $\cV$, and the target predictor is the conditional law of a designated target position given a prefix of preceding positions. Within this setting we characterize when source data enables zero-shot or few-shot learning of the target predictor (\emph{fast adaptation}), when it instead hinders learning (\emph{slow adaptation}), and what lies between. Our contributions are as follows:
\begin{enumerate}
    \item \textbf{Module and circuit transportability} (\Cref{sec:circuit-TR}). \emph{Module-TR} is the atomic case: the target predictor is a single source-side mechanism with possibly reordered parents (\Cref{alg:AD-multi-with-delta}). \emph{Circuit-TR} composes several such mechanisms through the target's causal graph (\Cref{alg:AD-Delta-sequence}); module-TR is the special case $T_*=M+1$. Given the source and target diagrams and a discrepancy oracle (\Cref{def:discrepancies-oracle}), the target risk is $\cO\!\big(\tfrac{|\cI|\,|\cV|^{c+1}}{\epsilon^2 N} + \tfrac{(T_*-M-|\cI|)\,|\cV|^{c+1}}{\epsilon\, n}\big)$ (\Cref{thm:structure-informed-sequence}), where $|\cI|$ counts target positions with a source-side match -- interpolating between zero-shot ($|\cI|=T_*-M$) and target-only ($|\cI|=0$).
    \item \textbf{Few-shot adaptation without causal structure} (\Cref{sec:adaptation}). \emph{Circuit-AD} (\Cref{alg:agnostic-sequence}) drops the diagrams and oracle: it enumerates candidate predictors built from source data alone and selects on held-out target data. Its excess risk over circuit-TR is $\cO(\sqrt{T_*^3 \log T_* / n})$ (\Cref{thm:agnostic-sequential}), so the achievable rate is governed by the size $L$ of the smallest target circuit composable from source-learnable modules; $L = \cO(\sqrt[3]{n})$ separates fast and slow adaptation (\Cref{cor:fast-slow-threshold}).
    \item \textbf{Gradient-based relaxation} (\Cref{sec:optimization}). Symbolic circuit-AD is combinatorial in $T_*$. We propose a gradient-based architecture that learns the same structural objects circuit-TR uses ($\Phi$, $A^{\src}$, $A^\star$, $\prob$) by stochastic optimization, and verify in \Cref{sec:simulations} that it qualitatively reproduces circuit-AD across three regimes: fast adaptation with and without process supervision over intermediate positions, and slow adaptation when no source mechanism matches.
\end{enumerate}

\textbf{Preliminaries.} We use capital letters to denote variables ($X$), small letters for their values ($x$), bold letters for sets of variables ($\bX$) and their values ($\bx$), and use caligraphic letters ($\cX$) to denote their support. A conditional independence statement in distribution $P$ is written as $(\bX \indep \bY \mid \bZ)_P$. A $d$-separation statement in some graph $\cG$ is written as $(\bX \indep_d \bY \mid  \bZ)$. To denote $P(\bY = \by \mid \bX = \bx)$, we use the shorthand $P(\by\mid \bx)$.
The basic semantic framework of our analysis relies on Structural Causal Models (SCMs) \citep[Definition 7.1.1]{pearl2009causality}, which are defined below.

\begin{definition}
     \textit{An SCM $\cM$ is a tuple $M = \langle \bV , \bU, \cF, P \rangle$ where each observed variable $V \in \bV$ is a deterministic function of a subset of variables $\Pa_V \subset \bV$ and latent variables $\bU_V \subset \bU$, i.e., $v := f_V (\pa_V, \bu_V), f_V \in \cF$. The unobserved variables $\bU$ follow a distribution $P(\bu)$.} \hfill $\square$
\end{definition} 

We assume the model to be recursive, i.e. that there are no cyclic dependencies among the variables. SCM $\cM$ entails a probability distribution $P^{\cM}(\bv)$ over the set of observed variables $\cV$ such that 
\begin{equation} \label{eq:truncated-prod}
    P^\cM(\bv) = \int_{\cU} \prod_{V \in \bV} P^{\cM}( v\mid \pa_V, \bu_V)\cdot P(\bu) \cdot d\bu,
\end{equation}
where the term $P(v\mid \pa_V, \bu_V)$ corresponds to the function $f_V \in \cF$ in the underlying structural causal model $\cM$. It also induces a causal diagram $\cG_{\cM}$ in which each $V\in\bV$ is associated with a vertex, and we draw a directed edge between two variables $V_i \rightarrow V_j$ if $V_i$ appears as an argument of $f_{V_j}$ in the SCM, and a bi-directed edge $V_i \leftrightarrow V_j$ if $\bU_{V_i} \cap \bU_{V_j} \neq \emptyset$ or $P(\bU_{V_i}, \bU_{V_j}) \neq   P(\bU_{V_i})\cdot P(\bU_{V_j})$, that is $V_i$ and $V_j$ are confounded \citep{bareinboim:etal20}.

Throughout this paper, following standard practice in nonparametric causal identification \citep[e.g.,][]{pearl2009causality,bareinboim2014transportability,correa2019statistical}, we restrict to recursive SCMs over discrete variables with strict positivity ($P^{\cM}(\bv) > \epsilon$ for every $\bv$ and a known $\epsilon$) and no unobserved confounders, and operate non-parametrically; we revisit confounding in \Cref{app:confounders} and discuss the role of these assumptions in \Cref{app:limitations}.
\section{Circuit Transportability} \label{sec:circuit-TR}

A target SCM, viewed as an ordered list of probabilistic mechanisms acting on its observed variables, is a \emph{circuit}; its \emph{size} is the number of positions $T_*$. We say the query $P^\star(y\mid \bx)$ is \emph{circuit-transportable} from a source distribution when the target predictor admits an estimator built by composing mechanisms each drawn from some source position (formal in \Cref{def:circuit-TR}). This section develops the structure-informed regime, where the source and target causal graphs and a discrepancy oracle are given; \Cref{sec:adaptation} relaxes these assumptions in favor of a limited slice of labeled target data. We start from an atomic case (\emph{module-transportability}, where the target predictor is a single reused source mechanism) and lift it to circuit-transportability, where the target predictor is a sequential composition of such mechanisms wired through the target's causal graph.

\begin{wrapfigure}{r}{0.28\textwidth}
\centering
\begin{tikzpicture}[
    scale=0.65]
    \definecolor{domain1}{RGB}{31,119,180}
    \definecolor{domain2}{RGB}{255,127,14}
    \definecolor{domain3}{RGB}{44,160,44}
    
    \node[draw, circle, minimum size=0.7cm] (X1) at (-1.5,0) {$X_1$};
    \node[draw, circle, minimum size=0.7cm] (X2) at (0,0) {$X_2$};
    \node[draw, circle, minimum size=0.7cm] (X3) at (1.5,0) {$X_3$};
    
    \node[draw, circle, minimum size=0.7cm] (Y) at (0,-2) {$Y$};
    
    \node[draw, dashed, rectangle, fit={(X1) (X2) (X3)}, inner sep=0.2cm] (Xbox) {};
    \node at (-3,0.5) {$\mathbf{X}:$};
    
    \draw[->, thick, domain1] (X1) to[bend right=0] (Y);
    \draw[->, thick, double, domain1] (X2) to[bend right=20] (Y);
    
    \draw[->, thick,  double,domain2] (X2) to[bend left=20]  (Y);
    \draw[->, thick, domain2] (X3) to[bend left=0] (Y);
\end{tikzpicture}
\caption{Causal diagrams corresponding to \Cref{ex:structure-informed-multi}. Color-coded edges show parents of $Y$ in each domain: blue for $\mathcal{M}^{\src}$, orange for $\mathcal{M}^\star$. Single edges are the first parent and double edges are the second parent.}
\label{fig:multi-cause-example}
\vspace{-1cm}
\end{wrapfigure}

We consider a classification problem where $\bX = \{X_1,X_2,...,X_{M}\}$, and $Y$ are discrete-valued variables that take value in a shared finite vocabulary $\cV$. The objective is predicting the label $Y$ using covariates $\bX$, i.e., learning $P^\star(y\mid \bx_{1:M})$. There is a loss function $\ell(\mu;y,\bx)$, and the risk is defined as the expected loss $R_{P^\star}(\mu) := \bbE_{P^\star}[\ell(\mu;Y,\bX)]$. The true risk minimizer is denoted as $\mu_* \in \argmin_{\mu:\cV^{M_X}\to\mathrm{simplex}({|\cV|^{M_Y}})} R_{P^\star}(\mu)$, and the empirical risk minimizer w.r.t. data $D$ is denoted as,
\begin{equation}
    \hat{P}(y\mid \bx;D) \in \argmin_{\mu:\cV^{M_X}\to\mathrm{simplex}(|\cV|)} \sum_{y,\bx\in D} \ell(\mu;y,\bx).
\end{equation}
We consider the loss to be the negative log-likelihood $\ell(\mu;y,\bx) := -\log \mu(y\mid \bx)$ in this work, and the objective is to minimize the excess risk denoted by $R_{P^\star}(\mu) - R_{P^\star}(\mu_*)$. 

We have access to large source data drawn from a source domain $\cM^{\src}$ that entails the source distribution $P^{\src}(\bx,y)$, and our predictions are to be evaluated in the target domain $\cM^\star$ that entails $P^\star(\bx,y)$. We assume strictly positive mass for every combination of the variables, i.e., $P^{\src}(\bx,y), P^\star(\bx,y) > \epsilon$. The data is denoted by $D^{\src}, D^\star$ with $|D^{\src}| = N$ and $|D^\star| = n$, and typically $N \gg n$. Below is a simple instance of the problem.

\begin{example} [Motivating example]  \label{ex:structure-informed-multi}
\normalfont
Suppose $X_1,X_2,X_3,Y \in \{0,1,...,9\}$. The source domain $\cM^{\src}$ and the target domain $\cM^\star$ are described as follows:
\begin{align*}
    U_{X_1},U_{X_2},U_{X_3} &\sim P(u_{X_1},u_{X_2},u_{X_3}) \\
    U_Y &\sim \mathrm{Multinomial}(\mathrm{prob}:\{0.91,0.01,...,0.01\})\\
    X_m &\gets U_{X_m}, \quad \forall m\in \{1,2,3\}\\
    Y &\gets \begin{cases}
        X_1 - X_2 + U_Y\pmod{10} \quad \text{ in } \cM^{\src}\\
        X_3 - X_2 + U_Y\pmod{10} \quad \text{ in } \cM^\star\\
    \end{cases}
\end{align*}
The causal diagram corresponding to these SCMs is shown in \Cref{fig:multi-cause-example}. The causal parents of $Y$ are a different subset of covariates in the source and target, but the mechanism that decides $Y$ given these parents is shared between $\cM^{\src}$ and $\cM^\star$; it is a noisy subtraction of the second parent from the first.

Suppose we know the parents of $Y$ in both domains, i.e., we have access to the ordered sets $\Pa^{\src}_Y = \langle X_1,X_2\rangle$ and $\Pa^\star_Y = \langle X_3,X_2\rangle$. Moreover, suppose we know that the \textit{module} generating $Y$ is shared between the two domains, i.e., $f^\star_Y(a,b,u_Y) = f^{\src}_Y(a,b,u_Y)$ for all $a,b \in \{0,1,...,9\}$ and $P^\star(u_Y) = P^{\src}(u_Y)$. Using this information, we can train a \textit{modular predictor} for $Y$ from source data, $\mu_{\src}(y\mid a,b) = \hat{P}(y\mid X_1=a, X_2=b; D^{\src})$, and then plug the target parents into this predictor in the appropriate order. With large source data, no target data is needed (zero-shot generalization) once we have such qualitative knowledge.

Note that $P^{\src}(Y \mid \bS) \neq P^\star(Y \mid \bS)$ for any subset $\bS \subset \{X_1,X_2,X_3\}$: no distributional invariance holds. Once we unfold $\bX$ into $\{X_1,X_2,X_3\}$, the structure of the parents of $Y$ in each domain plus mechanism sharing enables transport. \hfill $\square$
\end{example}

To encode mechanism sharing between the source and target, we use the following notation, adapted from \citet{correa2019statistical,bareinboim2014transportability}.

\begin{definition} [Domain discrepancy set] \label{def:domain-discrepancies}
    The discrepancy set $\Delta \subseteq \bV$ contains a variable $V$ if there is a possible mismatch between the causal mechanism of $V$ in $\cM^{\src}$ and $\cM^\star$, i.e., either $f^{\src}_V \neq f^\star_V$ or $P^{\src}(\bu_V) \neq P^\star(\bu_V)$. \hfill $\square$
\end{definition}

\begin{wrapfigure}{R}{0.48\textwidth}
    \vspace{-0.5cm}
    \begin{minipage}{0.48\textwidth}
      \begin{algorithm}[H]
        \caption{Module-TR} \label{alg:AD-multi-with-delta}
        \begin{algorithmic}[1]
    \REQUIRE $D^{\src}, D^\star; \Delta, \Pa^{\src}_Y, \Pa^\star_Y$
    \ENSURE $\mu_{\TR}(y\mid \bx) \approx P^\star(y \mid \bx)$
    \STATE $c \gets |\Pa_Y^\star|$
    \IF{$Y \notin \Delta$}
      \STATE $D^{\TR}_\bR \gets D^{\src}[Y,\bR_{1:c} = \pa^{\src}_Y[1:c]] \cup D^\star[Y,\bR_{1:c} = \pa^\star_Y[1:c]]$
    \ELSE
      \STATE $D^{\TR}_\bR \gets D^\star[Y,\bR_{1:c} = \pa^\star_Y[1:c]]$
    \ENDIF
    \STATE $\mu(y\mid \br) \gets \hat{P}(y \mid \br_{1:c}; D^{\TR}_\bR)$
    \STATE \textbf{Return} $\mu_{\TR} \gets \mu(y\mid \br = \pa^\star_Y)$
    \end{algorithmic}
      \end{algorithm}
    \end{minipage}
    \vspace{-0.5cm}
\end{wrapfigure}

Notably, because the parents of $Y$ are different between the domains, the existing notions of transportability (e.g., \citet{correa2019statistical,lee2020generalized}) do not license transport in \Cref{ex:structure-informed-multi}.

Module-TR (\Cref{alg:AD-multi-with-delta}) generalizes the approach in \Cref{ex:structure-informed-multi}: if the $Y$ mechanism is shared between source and target ($Y \notin \Delta$), we pool source and target data, reorder parents to a matching scope, and fit a single predictor; otherwise we fall back to target-only ERM.

\begin{proposition} [Module-TR] \label{lem:structure-informed-multi} In \Cref{alg:AD-multi-with-delta} with high probability, 
\begin{equation}
    R_{P^\star}(\mu_{\TR}) - R_{P^\star}(\mu_*) =\begin{cases}
         \cO(\frac{|\cX|^c\cdot |\cY|}{\epsilon^2 \cdot N}) \quad \text{ if } Y \notin \Delta \\
         \cO(\frac{|\cX|^c\cdot |\cY|}{\epsilon \cdot n}) \quad \text{ otherwise }
    \end{cases}
\end{equation}
where $c = |\Pa^\star_Y| \leq M$. \hfill $\square$
\end{proposition}

All proofs are in \Cref{app:proofs}. If $Y \in \Delta$, the source data does not help learning in the target: the target mechanism of $Y$ may be \textit{any} function, and the best error rate decays with $n$ as it would for ERM on $D^\star$. If $Y \notin \Delta$, the mechanism is shared, and knowledge of the discrepancy plus the parent sets lets us pool source and target data after reordering covariates to match the parents of $Y$ in the target. The resulting predictor uses up to $N+n$ effective samples, where $N \gg n$ — this is the zero-shot regime, with rate dominated by source size.

\begin{figure}
  \centering

  \tikzset{
    node/.style={circle, draw, minimum size=0.5cm, font=\small},
    box/.style={rectangle, draw, dashed, inner sep=6pt, minimum height=1.2cm},
    arrow/.style={->, >=stealth, thick}
}
  \begin{subfigure}{0.22\textwidth}
    \centering
    \begin{tikzpicture}[
      vertex/.style={circle, draw, minimum size=0.7cm, inner sep=1pt, font=\small},
      edge label/.style={font=\small}
    ]

    \node[vertex] (V1) at (0,0) {$X_1$};
    \node[vertex] (V2) at (1.3,0) {$X_2$};
    \node[vertex] (V3) at (2.6,0) {$Y_i$};
    
    \draw[->, thick] (V1) to[bend left=30] node[edge label , near end, above] {$\max,\min,-$} (V3);
    \draw[->, thick] (V2) to (V3);
    
    \end{tikzpicture}
    \caption{Three positions in $\cG^{\src}$}
    \label{fig:sequential-example-source}
  \end{subfigure}%
  \begin{subfigure}{0.75\textwidth}
    \centering
    \begin{tikzpicture}[
      vertex/.style={circle, draw, minimum size=0.7cm, inner sep=1pt, font=\small},
      edge label/.style={font=\small}
    ]
    
    \node[vertex] (V1) at (0,0) {$V_1$};
    \node[vertex] (V2) at (1.3,0) {$V_2$};
    \node[vertex] (V3) at (2.6,0) {$V_3$};
    \node[vertex] (V4) at (3.9,0) {$V_4$};
    \node[vertex] (V5) at (5.2,0) {$V_5$};
    \node (Vdots1) at (6.5,0) {$\cdots$};
    \node[vertex] (V6) at (7.7,0) {$V_{3|\cV|}$};

    \node[box, fit={(V3) (V4) (V5)}] (Zbox) {};
    \node at (3.7,0.9) {repeat $|\cV|$ times};

    \node[box, fit={ (V1) (V2)}] (Xbox) {};
    \node at (0.6,0.9) {$\bX = \{X_1,X_2\}$};

    \node[box, fit={ (V6)}] (Ybox) {};
    \node at (7.7,0.9) {$Y$};
    
    \draw[->, thick] (V2) to[bend left=0] node[edge label , near end, below] {$\max$} (V3);
    \draw[->, thick] (V1) to[bend left=30] (V3);
    \draw[->, thick] (V3) to[bend left=0] node[edge label, near end, below] {$\min$} (V4);
    \draw[->, thick] (V2) to[bend left=30] (V4);
    \draw[->, thick] (V3) to[bend left=30] (V5);
    \draw[->, thick] (V4) to[bend left=0]node[edge label, near end, below] {$-$} (V5);
    \draw[->, thick] (V5) to[bend left=0] (Vdots1);
    \draw[->, thick] (Vdots1) to[bend left=0] (V6);
    \end{tikzpicture}
    \caption{$\mathcal{G}^\star$}
    \label{fig:sequential-example-target}
  \end{subfigure}
  \caption{Causal graphs corresponding to \Cref{ex:beyond-module-TR}. The source contains three positions whose mechanisms are noisy $\max$, $\min$, and $-$ on a pair of independent uniform inputs $V^{\src}_1, V^{\src}_2$ (left); the target $\tilde{\cM}^\star$ implements a GCD algorithm by repeating the same three operators through a chained structure (right). The query of interest is $P^\star(v^\star_{3|\cV|}\mid v^\star_1,v^\star_2)$.}
  \label{fig:sequential-example}
\end{figure}

\begin{example} [Beyond module-transportability]\label{ex:beyond-module-TR}
\normalfont
Consider a source SCM $\cM^{\src}$ over $T=5$ positions and a target SCM $\cM^\star$ over $T_*=3$ positions, sharing the discrete vocabulary $\cV$. The first two positions of each domain are independent uniform inputs, $V^{\src}_1, V^{\src}_2 \sim \mathrm{unif}(\cV)$ and $V^\star_1, V^\star_2 \sim \mathrm{unif}(\cV)$. The remaining source positions implement noisy operators on $V^{\src}_1, V^{\src}_2$:
\begin{align*}
    V^{\src}_3 &\gets \max(V^{\src}_1,V^{\src}_2) + U_3, & V^{\src}_4 &\gets \min(V^{\src}_1,V^{\src}_2) + U_4, & V^{\src}_5 &\gets V^{\src}_1 - V^{\src}_2 + U_5.
\end{align*}
The target's third position is a noisy GCD: $V^\star_3 \gets f^\star_3(V^\star_1,V^\star_2,U_3) \approx \mathrm{GCD}(V^\star_1,V^\star_2)$, and $Y := V^\star_3$. We require the operators to be noisy to avoid violating positivity (cf.\ Preliminaries). Module-TR (\Cref{alg:AD-multi-with-delta}) cannot transport $P^\star(y\mid v^\star_1, v^\star_2)$ from the source: no single source position implements GCD.

A noisy GCD can, however, be \textit{composed} from the source-learnable modules. Consider an alternative target SCM $\tilde{\cM}^\star$ over $T_*=3|\cV|$ positions with $V^\star_1,V^\star_2 \sim \mathrm{unif}(\cV)$ as before and, for $i\in\{1,2,\dots,|\cV|\}$,
\begin{align}
    V^\star_{3i} \gets \max(V^\star_{3i-1},V^\star_{3i-2}),\quad V^\star_{3i+1} \gets \min(V^\star_{3i-1},V^\star_{3i-2}), \quad V^\star_{3i+2} \gets V^\star_{3i} - V^\star_{3i+1}.
\end{align}
This SCM implements GCD via the $\min,\max,-$ operators, i.e.,
\begin{equation}
    P^{\tilde{\cM}^\star}(v^\star_{3|\cV|} \mid v^\star_1,v^\star_2) \approx \mathbf{1}\{v^\star_{3|\cV|} = \mathrm{GCD}(v^\star_1,v^\star_2)\}.
\end{equation}
With domain knowledge encoding this compositional structure and the shared mechanisms with $\cM^{\src}$, we can learn each base operator from the appropriate source position and compose them to obtain a noisy GCD predictor. The query $P^\star(y\mid v^\star_1, v^\star_2)$ is thus transportable, but only under a richer notion of domain knowledge. \hfill $\square$
\end{example}

\Cref{ex:beyond-module-TR} motivates a more general approach for transportability when the target query arises through compositional structure.

Let $\bV^{\src} = \{V^{\src}_1,\dots,V^{\src}_{T}\}$ and $\bV^\star = \{V^\star_1,\dots,V^\star_{T_*}\}$ denote the observable variables in $\cM^{\src}$ and $\cM^\star$, respectively. Each variable takes value in a finite set $\cV$, shared across positions and domains. We assume a causal order in each SCM and no unobserved confounding. Let $\cG^{\src} = \{\Pa^{\src}_i\}_{i=1}^{T}$ and $\cG^\star = \{\Pa^\star_i\}_{i=1}^{T_*}$ be the causal diagrams (e.g., \Cref{fig:sequential-example}). The label is $Y := V^\star_{T_*}$ and the covariates are $\bX := \{V^\star_1,\dots,V^\star_M\}$. The goal is to learn $P^\star(y\mid \bx) = P^\star(v^\star_{T_*}\mid \bv^\star_{1:M})$ using $N$ source samples and $n$ target samples. The special case $T_* = M+1$ coincides with the module-transportability task.

As seen in \Cref{ex:beyond-module-TR}, the discrepancy structure in the sequential setting may identify mechanisms across different positions of source and target. Below is an extension of the discrepancy notion that accommodates such commonalities.

\begin{definition} [Discrepancy oracle] \label{def:discrepancies-oracle} Let $\Delta(i, i')$ be a boolean function over target position $i \in [T_*]$ and source position $i' \in [T]$ that returns one if either $f^\star_i \neq f^{\src}_{i'}$ or $P^\star(u^\star_i) \neq P^{\src}(u^{\src}_{i'})$, and returns zero otherwise. \hfill $\square$
\end{definition}

We treat $\Delta$ and $\{\cG^{\src}, \cG^\star\}$ as inputs throughout, but read them as a \emph{structured prior} over mechanism reuse rather than as ground truth: the agnostic adaptation in \Cref{sec:adaptation} searches over candidate $\Delta$, and our experiments quantify graceful degradation under partial corruption of $\Delta$. We can now formally define circuit transportability.

\begin{definition} \label{def:circuit-TR} The query $P^\star(v^\star_{T_*}\mid \bv^\star_{1:M})$ is circuit-transportable from the source distribution $P^{\src}$ given the discrepancy oracle $\Delta$ and the causal diagrams $\cG^{\src}, \cG^\star$, if for every pair of source and target SCMs $\cM^{\src}, \cM^\star$ that induce these diagrams and $\Delta$ and entail $P^{\src}$, the conditional $P^{\cM^\star}(v^\star_{T_*}\mid \bv^\star_{1:M})$ is uniquely determined.
\end{definition}

\begin{wrapfigure}{R}{0.59\textwidth}
\vspace{-1cm}
    \begin{minipage}{0.59\textwidth}
      \begin{algorithm}[H]
        \caption{Circuit-TR} \label{alg:AD-Delta-sequence}
        \begin{algorithmic}[1]
    \REQUIRE $D^{\src}, D^\star$; $\Delta$; $\cG^{\src}, \cG^\star$
    \ENSURE $\mu_{\TR}(v^\star_{T_*}\mid v^\star_{1:M}) \approx P^\star(v^\star_{T_*}\mid v^\star_{1:M})$
    \FOR{ $i \in \{M+1,\dots,T_*\}$}
    \STATE $\cJ_i \gets \{i' \in [T] : \Delta(i, i') = 0\}$
    \STATE $D^{\TR}_i \gets \bigcup_{i' \in \cJ_i} D^{\src}[Y\!:\!V^{\src}_{i'},\, \bX_{1:c}\!:\!\Pa^{\src}_{i'}] \,\cup\, D^\star[Y\!:\!V^\star_i,\, \bX_{1:c}\!:\!\Pa^\star_i]$
    \ENDFOR
    \STATE $\mu_{\TR} = \prod_{i=M+1}^{T_{*}} \hat{P}(Y=v^\star_i\mid \bX = \pa^\star_i;\, D^{\TR}_i)$
    \STATE \textbf{Return} $\mu_{\TR}(v^\star_{T_*}\mid v^\star_{1:M}) \gets \sum_{v^\star_{M+1:T_*-1}} \mu_{\TR}$
    \end{algorithmic}
      \end{algorithm}
    \end{minipage}
    \vspace{-0.cm}
\end{wrapfigure}

In the context of \Cref{ex:beyond-module-TR}, the source positions $V^{\src}_3, V^{\src}_4, V^{\src}_5$ implement noisy $\max, \min, -$ respectively. The target's $V^\star_{3i}$ position (which uses $\max$) shares mechanism with source position $3$, so $\Delta(3i, 3) = 0$; similarly $\Delta(3i+1, 4) = \Delta(3i+2, 5) = 0$. The target's GCD output position $Y$ does not share mechanism with any single source position.

Circuit-TR (\Cref{alg:AD-Delta-sequence}) is a general procedure for the circuit-transportability task. The discrepancy oracle $\Delta$ and the causal diagrams $\cG^{\src}, \cG^\star$ identify, for each target position, the source positions sharing its mechanism; the pooled data trains the corresponding conditional $P^\star(v^\star_i\mid \pa^\star_i)$. These conditionals are composed to yield an estimator of $P^\star(v^\star_{M+1:T_*} \mid v^\star_{1:M})$, and the auxiliary variables $V^\star_{M+1},\dots,V^\star_{T_*-1}$ are marginalized out to obtain $P^\star(v^\star_{T_*}\mid v^\star_{1:M})$. 

Remark that module-TR achieves either zero-shot generalization (rate in $N$) or target-only ERM (rate in $n$) (\Cref{lem:structure-informed-multi}). Circuit-TR (\Cref{alg:AD-Delta-sequence}) may lie between these extremes: if every target position's mechanism matches some source position, all modules transport and the rate depends on $N$; if no target position has a match, the rate depends on $n$; in between, the bound has fast and slow components in $N$ and $n$, respectively. Below is an upper bound for circuit-TR's target risk.

\begin{theorem} [Circuit-TR error rate] \label{thm:structure-informed-sequence} Let $\cI = \{i \in \{M+1,\ldots,T_*\}: \cJ_i \neq \emptyset\}$ index the target positions whose mechanism transports, and let $c = \max_i |\Pa^\star_i|$. In \Cref{alg:AD-Delta-sequence}, with high probability,
\begin{equation}
    R_{P^\star}(\mu_{\TR}) - R_{P^\star}(\mu_*) = \cO\!\Big(\tfrac{|\cI|\,|\cV|^{c+1}}{\epsilon^2 N} \,+\, \tfrac{(T_*-M-|\cI|)\,|\cV|^{c+1}}{\epsilon \, n}\Big).
\end{equation}
The two extremes recover the familiar dichotomy: when every target position transports ($|\cI| = T_*-M$) the rate decays in $N$ alone; when none transports ($|\cI| = 0$) the rate decays in $n$ alone. The intermediate spectrum is discussed in \Cref{app:structure-informed-rates}.
\end{theorem}

A transportable circuit can be read as a \emph{causal/mechanistic interpolation} from the source: every target mechanism $f^\star_i$ must appear at some source position. When at least one target position has no source match, the rate of \Cref{thm:structure-informed-sequence} carries a term decaying in $n$ (slow adaptation).

Such structural knowledge is rarely available in practice, but a small amount of labeled target data often is. The next section asks whether circuit-transportability still helps when $\Delta, \cG^{\src}, \cG^\star$ are unknown and only the target sample $D^\star$ is given.

\section{Few-shot Learning via Circuit Transportability} \label{sec:adaptation}

\begin{wrapfigure}{R}{0.61\textwidth}
\vspace{-0.8cm}
    \begin{minipage}{0.61\textwidth}
      \begin{algorithm}[H]
        \caption{Circuit-AD} \label{alg:agnostic-sequence}
        \begin{algorithmic}[1]
    \REQUIRE $D^{\src}, D^\star$
    \ENSURE Target classifier $\hat{\bbE}_{P^\star}[Y \mid \bx]$
    \STATE Let $\cE = [T] \cup \{1^\star,\dots,T_*^\star\}$ (source and target positions), $\cH \gets \emptyset$
    \STATE $D^\star_{\mathrm{tr}}, D^\star_{\mathrm{te}} \gets \mathrm{partition}(D^\star)$
    \FOR{ every partition of $\cE$ into clusters $\cS = \{\cE_l\}_l$}
    \STATE $\Delta(i, i') \gets \begin{cases}
        0 & \text{if } \exists \cE_l \in \cS \text{ with } \{i^\star, i'\} \subseteq \cE_l\\
        1 & \text{otherwise}
    \end{cases}$
    \FOR{ every pair of graphs $(\cG^{\src}, \cG^\star)$}
    \STATE $\cH \gets \cH \cup \{\mathrm{circuitTR}(D^{\src}, D^\star_{\mathrm{tr}};\, \Delta;\, \cG^{\src}, \cG^\star)\}$
    \ENDFOR
    \ENDFOR
    \STATE \textbf{Return} $\mu_\AD \gets \argmin_{\mu \in \cH} \ell(\mu; D^\star_{\mathrm{te}})$
    \end{algorithmic}
      \end{algorithm}
    \end{minipage}
    \vspace{-0.5cm}
\end{wrapfigure}

Suppose we do not have access to a domain knowledge comprised of a collection of causal graphs and the discrepancy oracle. The next example depicts a situation where we can still make accurate predictions in the target domain.

\begin{example} \label{ex:agnostic-multi}
\normalfont
In the context of \Cref{ex:structure-informed-multi}, suppose we do not have access to the discrepancy set $\Delta$ and the parent sets $\Pa^{\src}_Y,\Pa^\star_Y$. To compensate for this lack of knowledge, we take the following approach: we train a collection of predictors that would contain at least one candidate whose risk in the target is as good as what is achievable through module-TR, and then we use held-out target data to find the best performing one among these candidates, hopefully identifying the best-performing one with few target samples. In particular, for every $c\in \{0,1,2,3\}$, we take two ordered set of the covariates of size $c$, and regress $Y$ on these $c$ variables in the specified order, once using target data alone and once using source and target data combined. this results in a total of at most $\sum_{c=0}^3 ({3 \choose c}\cdot c!)^2 + {3 \choose c}\cdot c! = 98$ predictors. Notice that this number does not depend on dimensionality of $X,Y$, and is only a function of the number of variables. Finally, we use held-out target data to choose the best performing one in this pool of predictors. This imposes a fixed excess risk on top of what is achievable through module-TR, and since we know that this example is in fact module-transportable, we can guarantee a small total risk with only a few target samples. \hfill $\square$
\end{example}

As seen in the example above, each hypothetical domain knowledge yields an estimator for $P^\star(v^\star_{T_*}\mid v^\star_{1:M})$ via circuit-TR. Since the discrepancy oracle $\Delta$ and the diagrams $\cG^{\src}, \cG^\star$ are discrete objects, fixing $T_*$ leaves finitely many candidate estimators. Partition target data into training and validation sets of size $\tfrac{n}{2}$, build candidates using the training half plus the source data, and use the validation half to select the best performer. Circuit-AD (\Cref{alg:agnostic-sequence}) implements this. What follows is its performance guarantee.

\begin{theorem} [Circuit-adaptation rate] \label{thm:agnostic-sequential}
Let $\mu_{\TR},\mu_\AD$ be learned by circuit-TR (\Cref{alg:AD-Delta-sequence}) and circuit-AD (\Cref{alg:agnostic-sequence}), respectively. Then,
\begin{equation} \label{eq:rate-agnostic-sequence}
    R_{P^\star}(\mu_\AD) = \cO\!\Big(R_{P^\star}(\mu_{\TR}) + \sqrt{\tfrac{T_*^3 \log T_*}{n}}\Big)
\end{equation}
where $T_*$ is the size of the target SCM.
\end{theorem}

Circuit-AD's guarantee is thus only marginally worse than circuit-TR's, despite the absence of explicit structural knowledge. An important consideration is the choice of $T_*$ -- the size of the target circuit. The example below emphasizes the subtleties of this choice.

\begin{example}[Slow rate for GCD]
\normalfont
Recall \Cref{ex:beyond-module-TR}, where GCD deems circuit-transportable from $\max,\min,-$, given, e.g., the causal graph in \Cref{fig:sequential-example-target}. Now suppose we do not have access to the causal graphs and domain discrepancies, so we need to use circuit-AD. We need to make a choice of $T_*$, which circuit-AD takes as input and does not prescribe. In hindsight, we can ensure that $T_* = 3|\cV|$ is large enough, since there exists a recipe of this length (i.e., causal graph and discrepancies oracle) that licenses circuit-transportability. Due to \Cref{thm:agnostic-sequential}, this choice offers the following guarantee:
\begin{equation}
    R_{P^\star}(\mu_\AD) = \cO(R_{P^\star}(\mu_{\TR}) + \sqrt{\frac{ |\cV|^3\cdot \log |\cV|}{n}}). \label{eq:bad-excess-risk}
\end{equation}
This is a slow rate, since $n = \Omega(|\cV|^3)$ target samples are needed for constant excess risk; vanilla ERM on the target yields a similar rate. \hfill $\square$
\end{example}
\begin{center}
    \emph{Which set of modules should the target domain have for $P^\star(y\mid \bx)$ to be circuit-transportable?}
\end{center}

This is known as \textit{Minimum Circuit Size Problem (MCSP)} \citep{
Shannon1949,Lupanov1958,FurstSaxeSipserFOCS1981,AroraBarak2009,MurrayWilliams2017,AllenderGrochowVanMelkebeekMooreMorgan2018}. The modules considered in this circuit complexity theory are often deterministic boolean functions, but, there are also probabilistic extensions too, cf. \citet{AdlemanFOCS1978,YaoFOCS1977,HastadSTOC1986}. \textit{Clone membership problem (CMP)}  \citep{Post1941,Lau2006,DBLP:journals/mst/Vollmer09} concerns whether it is even possible to transport a circuit from a fixed set of modules. CMP is indeed \textit{decidable} in time and space polynomial in size of the conditional distribution table (exponential in vocabulary size $|\cV|$). If the source-learnable modules form a \emph{functionally complete basis}, every conditional distribution is circuit-transportable at \emph{some} target circuit size; but adaptation rates still depend on that size, so functional completeness alone does not yield fast adaptation. The example below illustrates how a circuit's adaptability depends on which modules the source provides.

\begin{example} [Fast rate for GCD]
\normalfont
In the context of \Cref{ex:beyond-module-TR}, suppose the source additionally contains a position $V^{\src}_6$ implementing $a \bmod b$. An algorithm for GCD using $\max,\min,\mathrm{mod}$ replaces the $-$ operator with $\mathrm{mod}$, reducing the target circuit size to $T_* = 3\lceil\log_2|\cV|\rceil$. Circuit-AD then enjoys a much better excess risk than \Cref{eq:bad-excess-risk}, though the guarantee still depends polylogarithmically on the vocabulary size. \hfill $\square$
\end{example}

Motivated by this observation, we have the following statement.
\begin{corollary}[Fast/slow] \label{cor:fast-slow-threshold}
Let $L$ be the minimum size of a target circuit composable from source-learnable modules that computes $P^\star(y\mid \bx)$. If $L$ is constant, circuit-AD achieves very fast adaptation rates. The threshold $L = \cO(\sqrt[3]{n})$ separates fast and slow adaptation regimes for a fixed target dataset of size $n$.
\end{corollary}

\vspace{-0.5cm}
\section{A gradient-based relaxation of circuit-AD} \label{sec:optimization}

The rate guarantees in \Cref{thm:agnostic-sequential,thm:structure-informed-sequence} are stated for the symbolic circuit-AD (\Cref{alg:agnostic-sequence}), whose enumeration over partitions of $[T]\cup[T_*]$ and pairs of causal diagrams is combinatorial in $T_*$. We propose a gradient-based relaxation to circuit-TR.

\subsection{The architecture and pretraining} \label{sec:pre-training}

We discuss the case $|\Pa^{\src}_i| \leq 1$ for all $i\in [T]$, removing this restriction in \Cref{app:multiple-parents-sequence}. Pretraining uses the source data $D^{\src}$ to learn mappings satisfying:
\begin{enumerate}[nosep]
    \item Mechanism indicator $\Phi: [T] \to [d]$ such that
    \begin{equation} \label{eq:condition-mechanisms}
        \Phi(i) = \Phi(i') \text{ iff } f^{\src}_i = f^{\src}_{i'} \text{ and } P^{\src}(u^{\src}_i) = P^{\src}(u^{\src}_{i'}).
    \end{equation}
    \item Parent matrix $A^{\src} \in [0,1]^{T\times T}$, lower-triangular, such that
    \begin{equation} \label{eq:condition-graphs}
        A^{\src}_{i,i'} = 1 \text{ iff } \Pa^{\src}_i = V^{\src}_{i'}.
    \end{equation}
    \item Universal predictor $\prob:\cV\times\cV\times[d] \to [0,1]$:
    \begin{equation} \label{eq:condition-prob}
        \forall y,x\in \cV,\ i\in [T]: \prob(y\mid x;\phi=\Phi(i)) = P^{\src}(V^{\src}_i = y\mid \Pa^{\src}_i = x).
    \end{equation}
\end{enumerate}
Intuitively, $\Phi$ clusters source positions by mechanism, $A^{\src}$ encodes the source causal diagram, and $\prob$ is the shared predictor for each cluster. Once \Cref{eq:condition-mechanisms,eq:condition-graphs,eq:condition-prob} hold, source prediction is optimal: $\hat{v}^{\src}_i \sim \prob(v^{\src}_i \mid X = A^{\src}_{i,\cdot}\cdot v^{\src}_{1:T}; \Phi(i))$. Any instantiation maximizing penalized source likelihood satisfies these properties.
\begin{proposition}  [Pretraining] \label{thm:pretraining} Let $\theta^\src$ be a parameterization such that
\begin{equation} \label{eq:pretraining}
    \theta^{\src} \in \argmin_{\theta \in \Theta} \,\bbE_{v^{\src}_{1:T}\sim P^{\src}}\!\Big[\sum_{i=1}^T -\log \prob_\theta\!\big(v^{\src}_i \mid {A_\theta^{\src}}_{i,\cdot}\cdot v^{\src}_{1:T}; \Phi_\theta(i)\big)\Big] + \lambda(d + \|A^{\src}_\theta\|_1),
\end{equation}
where $\Theta$ is the parameter space, $[d]$ is the range of $\Phi$, and $\|A^{\src}\|_1$ is the sum of entries of the parent matrix. Then $\Phi_{\theta^\src}, A^{\src}_{\theta^\src}, \prob_{\theta^\src}$ satisfy \Cref{eq:condition-mechanisms,eq:condition-graphs,eq:condition-prob}. \hfill $\square$
\end{proposition}

\begin{wrapfigure}{r}
{0.55\textwidth}
\vspace{-0.3cm}
    \centering
    \begin{subfigure}[b]{0.48\linewidth}
        \centering
        \includegraphics[width=\linewidth]{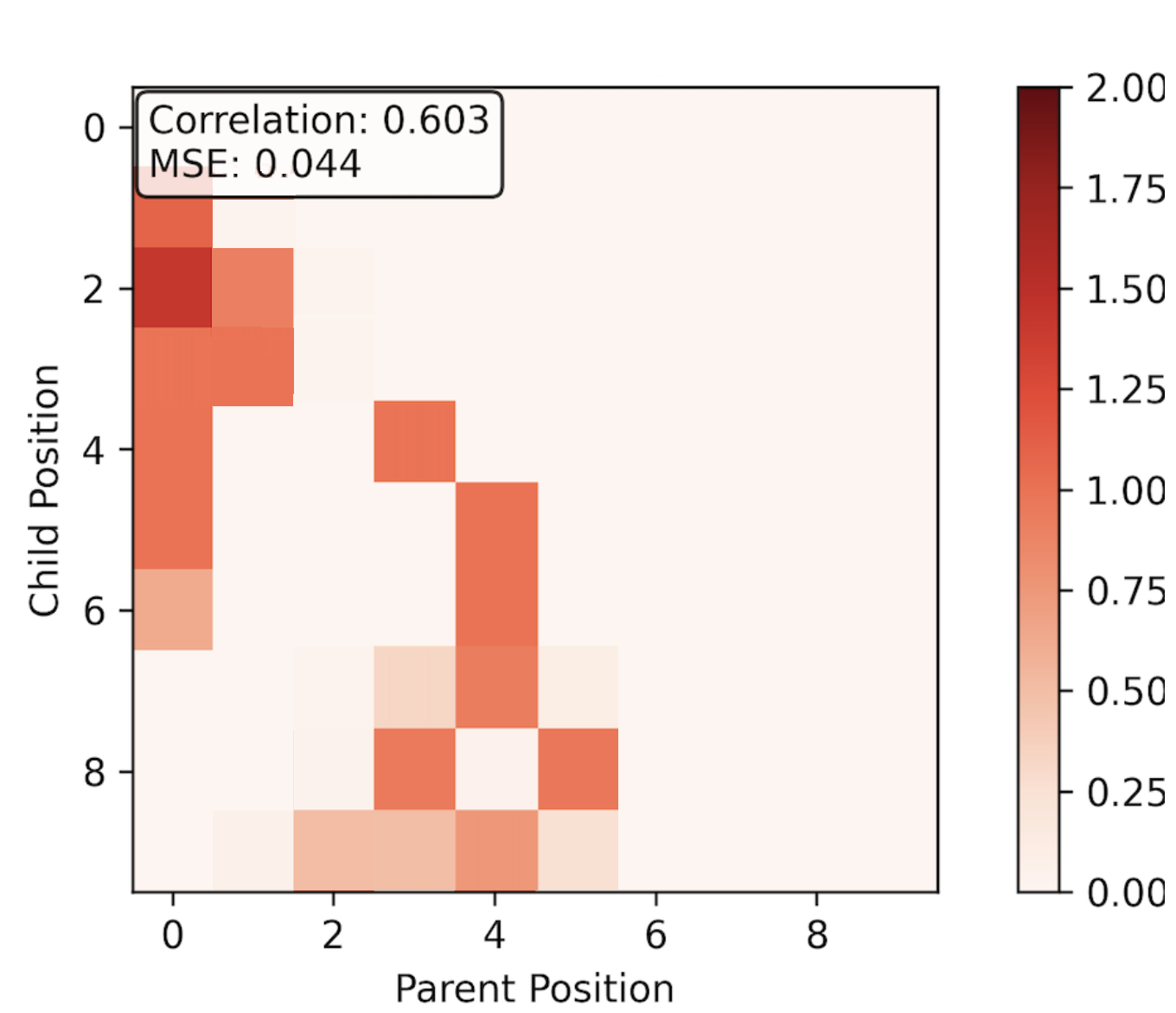}
        \caption{Learned graph}
    \end{subfigure}
    \hfill
    \begin{subfigure}[b]{0.48\linewidth}
        \centering
        \includegraphics[width=\linewidth]{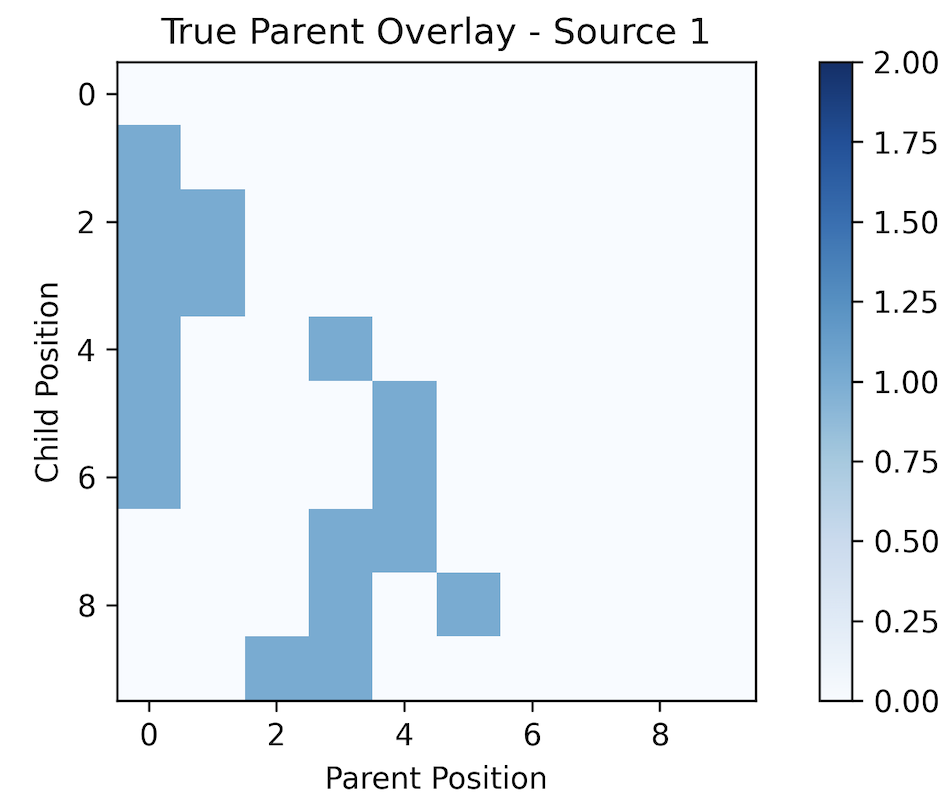}
        \caption{True graph}
    \end{subfigure}
    \caption{Implicit causal discovery in pretraining.}
    \label{fig:parent-matrices-over}
    \vspace{-1cm}
\end{wrapfigure}

As a sanity check on the pretraining objective, we run the relaxation on a synthetic source with $T\!=\!10$, $|\cV|\!=\!10$ and confirm that the learned $A^{\src}$ recovers the true source parent structure (\Cref{fig:parent-matrices-over}). A related phenomenon has been reported for transformers \citep{nichani2024transformers}; the additional element here is that the custom architecture also produces a within-source mechanism clustering through $\Phi$.

\subsection{Fine-tuning} \label{sec:fine-tuning}

Partition the target data $D^\star = D^\star_{\mathrm{tr}}\cup D^\star_\ft \cup D^\star_\mathrm{te}$ -- a refinement of the symbolic 2-way split in \Cref{alg:agnostic-sequence}: $D^\star_{\mathrm{tr}}$ fits the per-position target-only predictors, $D^\star_\ft$ learns the target-side parent attention and mechanism indicator, and $D^\star_\mathrm{te}$ trains the transport indicators (and corresponds to the held-out validation half of \Cref{alg:agnostic-sequence}). Once pretrained, each $\phi \in [d]$ corresponds to a cluster of source positions sharing a mechanism (\Cref{thm:pretraining}+\Cref{eq:condition-mechanisms}). Whenever $\Delta(i, i') = 0$ we have, for all $x,y\in\cV$, $P^\star(V^\star_i = y \mid \Pa^\star_i = x) = \prob_{\theta^\src}(V^{\src}_{i'} = y \mid X = x; \phi = \Phi(i'))$ via \Cref{eq:condition-prob}. Since $\prob_{\theta^\src}$ is fixed, we need only discover $\phi$ and $\Pa^\star_i$ at each target position $i$: introduce a target parent matrix $A^\star$ with $A^\star_{i,i'} = 1$ iff $\Pa^\star_i = V^\star_{i'}$ and a target mechanism indicator $\Phi^\star: [T_*] \to [d]$, and use $D^\star_\ft$ to learn
\begin{equation}
    \theta^{\trg} \in \argmin_{\theta \in \Theta} \sum_{v^\star_{1:T_*} \in D^\star_\ft} \sum_{i=1}^{T_*} - \log \prob_{\theta^\src}(Y = v^\star_i \mid X = {A_\theta^\star}_{i,\cdot}\cdot v^\star_{1:T_*}; \Phi_\theta^\star(i)).
\end{equation}

At each position $i\in [T_*]$ this is equivalent to using $|D^\star_\ft|$ target samples to pick the best predictor for $V^\star_i$ in $\cM^\star$ among $d\cdot(i-1)$ candidates from the sources. We additionally use $D^\star_{\mathrm{tr}}$ to fit a target-only model $\mu^\star_i := \hat{P}(v^\star_i \mid v^\star_{1:i-1};D^\star_{\mathrm{tr}})$ at each position, then learn \emph{transport indicators} $s^\star_i \in [0,1]$ that interpolate between the two:

\begin{equation}
    s^\star \in \argmin_{s\in [0,1]^{T_*}} \sum_{v^\star_{1:T_*} \in D^\star_\mathrm{te}} \sum_{i=1}^{T_*} -\log \!\big(\,s_i \cdot \prob_{\theta^\src}(v^\star_i \mid {A_{\theta^\trg}^\star}_{i,\cdot} \cdot v^\star_{1:T_*}; \Phi_{\theta^\trg}^\star(i)) + (1-s_i) \cdot \mu^\star_i(v^\star_i\mid v^\star_{1:i-1})\,\big).
\end{equation}
$s^\star_i \!\approx\! 0$ indicates target-only is best on held-out data and we forgo transport at $i$; $s^\star_i \!\approx\! 1$ indicates transport. We then estimate the query as

\vspace{-0.5cm}
\begin{equation*} \label{eq:adapt-final}
    \hat{\mu}_{\adapt}(v^\star_{T_*}\mid v^\star_{1:M}) = \sum_{v^\star_{M+1:T_*-1}} \prod_{i=M+1}^{T_*} \!\Big[\,s^\star_i \cdot \prob_{\theta^\src}(v^\star_i \mid {A_{\theta^\trg}^\star}_{i,\cdot} \cdot v^\star_{1:T_*}; \Phi_{\theta^\trg}^\star(i)) + (1-s^\star_i) \cdot \mu^\star_i(v^\star_i \mid v^\star_{1:i-1})\,\Big].
\end{equation*}

\begin{proposition} [Fine-tuning rate] \label{thm:fine-tuning-rate}
Let $\mu_{\AD}$ be learned by \Cref{alg:agnostic-sequence} and $\mu_\adapt$ (\Cref{eq:adapt-final}) be the result of the two-stage adaptation. Then $R_{P^\star}(\mu_\adapt) = \cO(R_{P^\star}(\mu_\AD))$.
\end{proposition}

\vspace{-0.5cm}
\subsection{Simulations} \label{sec:simulations}

We compare circuit-TR/AD against target-only ERM and two pooled baselines: \emph{ERM-pool} drops the domain indices, \emph{ERM-joint} preserves them. All methods share the same conditional-table estimator; the symbolic implementation realizes Algs.\,\ref{alg:AD-Delta-sequence}--\ref{alg:agnostic-sequence}. We use $|\cV|=10$, prefix $M=5$, and target SCMs drawn over the operator pool $\{\max,\min,-,\mathrm{add},\mathrm{add\text{-}one},\dots\}$. A chain has \emph{process supervision} when target data realises the intermediate positions $V^\star_{M+1},\dots,V^\star_{T_*-1}$, contributing per-step likelihood; when $T_*=M+1$ only the label is supervised. Three conditions: (a) \emph{fast adaptation, with process supervision}, $T_*=10$ (four intermediate target positions); (b) \emph{fast adaptation, without process supervision}, $T_*=6$ (no intermediates, same $\Delta$-search space as (a)); (c) \emph{slow adaptation}, $T_*=6$ with the target's last position implementing $x\cdot y \bmod |\cV|$, outside the source pool -- circuit-TR has no source match and the rate decays in $n$ (target-only ERM).

\begin{figure}[H]
    \centering
    \includegraphics[width=\linewidth]{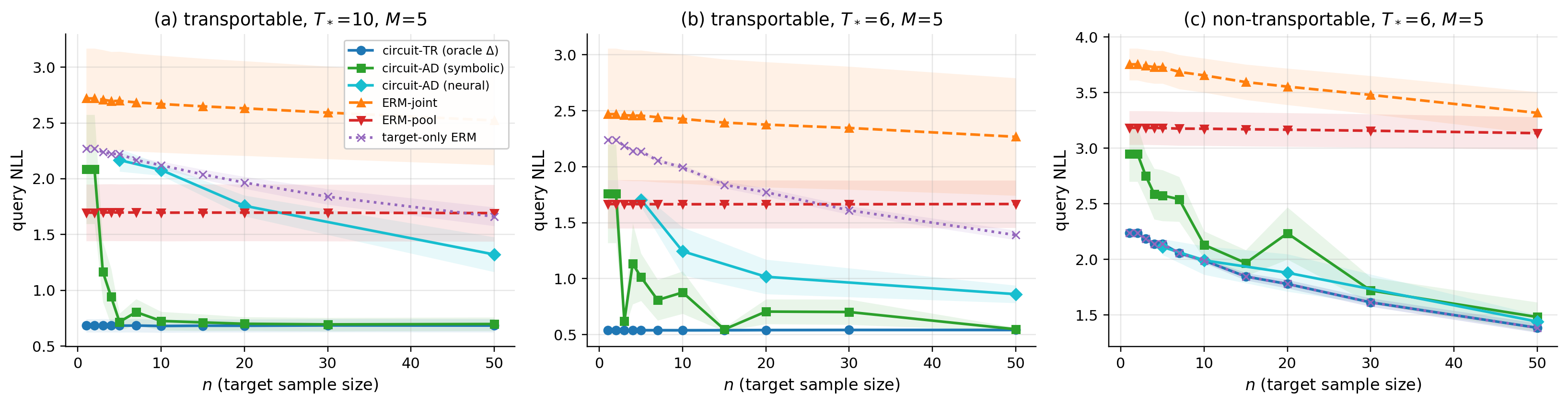}
    \caption{Few-shot adaptation ($n\!\leq\!50$). (a)~\emph{Fast adaptation, with process supervision} ($T_*\!=\!10$). (b)~\emph{Fast adaptation, without process supervision} ($T_*\!=\!6$). (c)~\emph{Slow adaptation} ($T_*\!=\!6$, last target operator outside the source pool). Full $n$-sweep in \Cref{fig:adapt-full}.}
    \label{fig:adapt-T=10}
\end{figure}

\Cref{fig:adapt-T=10} shows query-NLL on the target as $n$ grows from $1$ to $50$. With process supervision (a), circuit-AD locks onto oracle circuit-TR by $n\!\approx\!5$; without it (b), the same destination is reached by $n\!\approx\!15$. In (c), $\Delta_{\true}$ is identically $1$ for the query's row, so oracle circuit-TR coincides with target-only ERM and circuit-AD's search arrives at the same point; ERM-pool and ERM-joint sit $1$--$2$ nats higher because they pool source data that is wrong for the query position; more experiments in \Cref{app:experiments} (\Cref{fig:adapt-full,fig:scaling-misspec}); the orderings persist. Circuit-AD captures the full transport gain when the structure is there -- with or without process supervision -- and gracefully recovers target-only ERM otherwise, without being told which regime it is in.

\paragraph{Gradient-based relaxation.} The relaxation of \Cref{sec:pre-training,sec:optimization} (cyan, ``circuit-AD (neural)'' in \Cref{fig:adapt-T=10}) sits between symbolic circuit-AD and target-only ERM in (a) and (b) with a residual overhead of $0.1$--$0.3$ nats, and collapses onto target-only ERM in (c); the gradient-based search does not invent transport that is not there. We read this as qualitatively tracking the symbolic algorithm; tightening the gap is open (\Cref{app:limitations}).

\begin{figure}[H]
    \centering
    \includegraphics[width=0.6\linewidth]{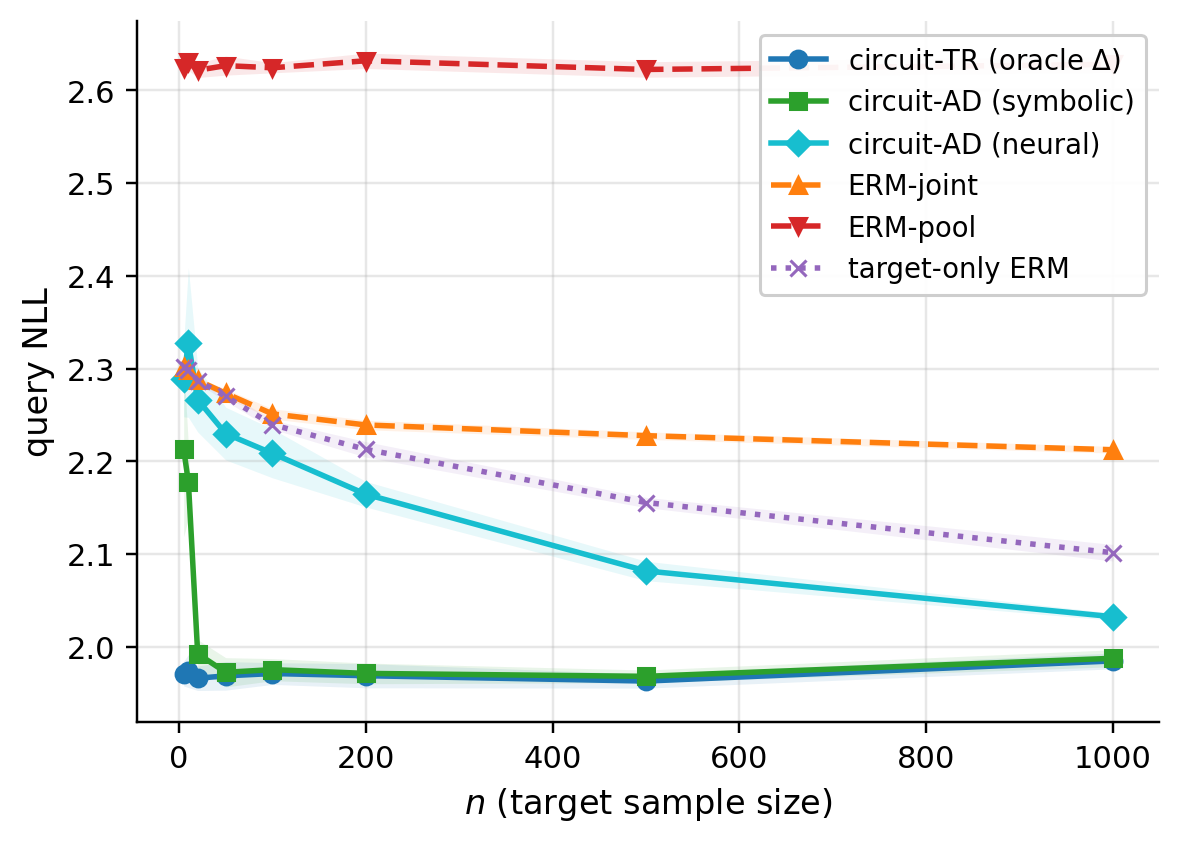}
    \caption{\emph{Held-out GCD composition}, $|\cV|\!=\!10$, $M\!=\!2$. Target is Euclidean's algorithm laid out as $(\max,\min,\bmod)\!\times\!5$ iterations plus a $\max$ readout, so $V^\star_{T_*-1}=\gcd(V^\star_0,V^\star_1)$ in the noiseless limit; source is an arithmetic-primitive library ($T\!=\!15$). Query NLL on $1000$ held-out target rows as $n$ grows from $5$ to $1000$. Circuit-AD matches oracle circuit-TR by $n\!\approx\!20$; target-only ERM is still $\approx\!0.13$ nats above the oracle at $n\!=\!1000$. Bands are mean $\pm$ 1\,SE over $5$ seeds (symbolic) and $3$ seeds (neural).}
    \label{fig:gcd-task}
\end{figure}

\paragraph{Real algorithmic-trace task: held-out GCD composition.} The conditions above use random source/target SCMs drawn from an operator pool. To check that the framework's gain is not an artifact of hand-constructed SCMs, we replace the target by a \emph{real} algorithm whose intermediate trace is dictated by the algorithm itself: Euclidean's GCD on $\cV=\{0,\dots,9\}$. The target SCM lays out five iterations of $(\max, \min, \bmod)$ followed by a $\max$ readout, so that $V^\star_{T_*-1}=\gcd(V^\star_0,V^\star_1)$ for every input pair under the noiseless limit. The source SCM exposes an arithmetic-primitive library ($T=15$ positions: $\mathrm{add}, \mathrm{subtract}, \mathrm{mult}, \min, \max, \bmod, \mathrm{copy}, \mathrm{add\text{-}one}, \mathrm{minus\text{-}one}$, with each of $\min, \max, \bmod$ appearing at two positions on different parent pairs); no single source position implements GCD, but every target mechanism appears at $\geq\!2$ source positions. We observe $M=2$ (the two operands) and predict $V^\star_{T_*-1}$. With $|\cV|=10$ this is exactly the held-out composition the reviewer flagged: the target's data-generating process is a real algorithm rather than a randomly drawn SCM, and circuit-AD must discover from a handful of target traces that the $(\max,\min,\bmod)$ modules transport from the source.
\Cref{fig:gcd-task} reports query NLL on $1000$ held-out target rows as $n$ grows from $5$ to $1000$. Symbolic circuit-AD locks onto oracle circuit-TR by $n\!\approx\!20$ and stays within $0.02$ nats of it through $n\!=\!1000$, while target-only ERM is still $\approx\!0.13$ nats above the oracle at $n\!=\!1000$ and ERM-joint plateaus $\approx\!0.22$ nats above the oracle. ERM-pool sits $\approx\!0.65$ nats above the oracle throughout: pooling every same-arity source position with every target position destroys the operator-specific signal at each step. The gradient-based relaxation (``circuit-AD (neural)'') tracks target-only ERM in the very-small-$n$ regime and then steadily improves, sitting roughly midway between symbolic circuit-AD and target-only ERM by $n\!=\!1000$ -- the same $\approx\!0.1$--$0.3$-nat overhead observed in \Cref{fig:adapt-T=10}.

This experiment exercises three properties at once: (i)~the target predictor is genuinely outside the source operator's single-position support, instantiating the compositional regime of Example~\ref{ex:beyond-module-TR}; (ii)~the ground-truth $\cG^\star$, $\cG^{\src}$, and $\Delta$ are determined by the algorithm itself rather than being inputs we fabricate, addressing the appendix-\ref{app:limitations} concern about synthetic-only evaluation; (iii)~the same circuit-AD code without modification produces near-oracle performance, which is the operational claim of \Cref{thm:agnostic-sequential}.

\vspace{-0.2cm}
\section{Conclusions}\vspace{-0.1cm}

\emph{Module-} and \emph{circuit-}transportability extend causal transportability theory to target predictors built by composing source-learnable mechanisms, licensing zero-shot prediction even when no source mechanism directly predicts the target label. Circuit-AD turns the recipe into a few-shot algorithm whose excess risk is governed by the \emph{size} of the smallest target circuit composable from source-learnable modules, not by total target complexity. The natural next bottlenecks are learning $\Delta$ and $(\cG^{\src},\cG^\star)$ from data, handling unobserved confounders (\Cref{app:confounders}), and tying the gradient relaxation in \Cref{sec:optimization} back to symbolic circuit-AD.

\begin{ack}
This research is supported in part by the NSF, ONR, AFOSR, DoE, Amazon, JP Morgan, and The
Alfred P. Sloan Foundation.
\end{ack}

\bibliography{bib}
\bibliographystyle{plainnat}

\newpage
\appendix
\DoToC

\newpage

\section{LLM usage disclosure} \label{app:llm-disclosure}
In this work, we used Chat-gpt and Claude, and Cursor for the following purposes:
\begin{itemize}
    \item Ideation, choice of notation, and structuring the paper.
    \item Literature overview and references.
    \item Developement of the codebase, and illustration/visualization.
    \item Content on reproducibility and implementation details in \Cref{sec:reproducibility}.
\end{itemize}

\section{Simplified uni-cause transportability and adaptation} \label{sec:uni-cause}

\begin{wrapfigure}{R}{0.45\textwidth}
    \begin{minipage}{0.45\textwidth}
      \begin{algorithm}[H]
        \caption{simple-AD ($X\to Y$)} \label{alg:AD-Delta}
        \begin{algorithmic}[1]
    \REQUIRE $D^{\src}, D^\star; \Delta$
    \ENSURE $\mu_{\TR} \approx P^\star(y\mid x)$
    \IF{$Y \notin \Delta$}
      \STATE $D^{\TR} \gets D^{\src} \cup D^\star$
    \ELSE
      \STATE $D^{\TR} \gets D^\star$
    \ENDIF
    \STATE \textbf{Return} $\mu_{\TR}\gets  \hat{P}(y\mid x;\, D^{\TR})$
    \end{algorithmic}
      \end{algorithm}
    \end{minipage}
\end{wrapfigure}

Consider the classification task where $Y$ takes value in a finite support $\cY$, and the covariate $X$ that takes value in a finite support $\cX$; in particular the objective is learning $P^\star(Y=y \mid X = x)$ within the hypothesis class containing all functions $\mu(y\mid x) : \cX \to \mathrm{simplex}^{|\cY|}$. There is a loss function $\ell(\mu;y,\bx)$, and the risk is defined as the expected loss $R_{P^\star}(\mu) := \bbE_{P^\star}[\ell(\mu;Y,\bX)]$. The true risk minimizer is denoted as $\mu_* \in \argmin_{\mu:\cX\to\mathrm{simplex}^{|\cY|}} R_{P^\star}(\mu)$, and the empirical risk minimizer w.r.t. data $D$ is denoted as,
\begin{equation}
    \hat{P}(y\mid \bx;D) \in \argmin_{\mu:\cX\to\mathrm{simplex}^{|\cY|}} \sum_{y,\bx\in D} \ell(\mu;y,\bx).
\end{equation}
We consider the loss to be the negative log-likelihood $\ell(\mu;y,x) := -\log \mu(y\mid \bx)$ in this work, and the objective is to minimize the excess risk denoted by $R_{P^\star}(\mu) - R_{P^\star}(\mu_*)$. 

Suppose we have target data $D^\star$ from a target domain $\cM^\star$ entailing $P^\star(x,y)$, and source data $D^{\src}$ from a source domain $\cM^{\src}$ entailing $P^{\src}(x,y)$. Let $n = |D^\star|$, $N = |D^{\src}|$, with $N \gg n$. We assume strict positivity, $P^{\src}(x,y), P^\star(x,y) > \epsilon$.

\begin{example} [Classification in $X\to Y$ case]  \label{ex:classification-no-confounding}
\normalfont
Suppose source $\cM^{\src}$ and target $\cM^\star$ are SCMs of the form
\begin{align*}
    U_X,U_Y &\sim \mathrm{unif}([0,1])\\
    X &\gets f_X(U_X)\\
    Y &\gets f_Y(X,U_Y),
\end{align*}
where $f_X, f_Y$ may differ between source and target. Both induce the diagram $X \to Y$, with $X$ the cause of $Y$ and no unobserved confounders. Without further assumptions the source data is unrelated to classification in the target; e.g., for $X \in \{0,1\}$ it is possible that $f^{\src}_Y = 1_{\{U_Y > 0.5\}}$ (so $Y \indep X$ in source) while $f^\star_Y(x,u_Y) = X \oplus 1_{\{U_Y > 0.9\}}$ (so $P^\star(Y=1\mid X=0) = 0.1$ and $P^\star(Y=1\mid X=1) = 0.9$). \hfill $\square$
\end{example}

 In the next example we use the domain discrepancy sets in an instance of the domain adaptation task.

\begin{example} [$\Delta$ might allow direct transport, or rule out any transport]  \label{ex:classification-no-confounding-TR}
\normalfont
In the context of \Cref{ex:classification-no-confounding}, suppose we have access to $\Delta$. If $Y \notin \Delta$, then $f^\star_Y = f^{\src}_Y$ and $P^\star(u_Y) = P^{\src}(u_Y)$, and
\begin{align*}
    P^\star(y \mid x) &= \sum_{u_Y} P^\star(y\mid u_Y,x)\, P^\star(u_Y\mid x) && \text{(introduce $U_Y$)}\\
    &= \sum_{u_Y} 1_{\{f^\star_Y(x,u_Y) = y\}}\, P^\star(u_Y) && \text{(defn.\ of $f^\star_Y$, $U_Y\indep X$)}\\
    &= \sum_{u_Y} 1_{\{f^{\src}_Y(x,u_Y) = y\}}\, P^{\src}(u_Y) = P^{\src}(y \mid x) && \text{($Y \notin \Delta$)}.
\end{align*}
So predicting in the target reduces to estimating $P^{\src}(y\mid x)$ from large source data. If $Y \in \Delta$, the discrepancy set rules out source-target mechanism sharing, and the source data is uninformative for $Y$ given $X$. \hfill $\square$
\end{example}

A simple procedure in \Cref{alg:AD-Delta} describes the above approach, and below is a guarantee.

\begin{lemma} [Simple-TR; $X\to Y$ case] \label{lem:simple-guarantees} In \Cref{alg:AD-Delta} with high probability,
\begin{equation}
    R_{P^\star}(\mu_{\TR}) - R_{P^\star}(\mu^\star)   = \begin{cases}
        \cO(\tfrac{|\cX|\cdot |\cY|}{\epsilon^2 \cdot N}) & \text{if } Y \notin \Delta\\
        \Omega(\tfrac{|\cX|\cdot |\cY|}{\epsilon \cdot n}) & \text{otherwise.}
    \end{cases}
\end{equation}
\hfill $\square$
\end{lemma}

If the source $Y$ mechanism matches the target's, support overlap (due to $P(x,y) \geq \epsilon$) lets simple-TR achieve fast rates depending on the large source size $N$. This is the \emph{transportable} (or zero-shot / domain-generalization) regime. Otherwise, even with $\Delta$, target and source SCMs admitting $\Delta$ may disagree on $P^\star(y\mid x)$ entirely, so no adaptation is possible. The dichotomy is a consequence of the discrete nature of $\Delta$ and the absence of confounding (no $U$ pointing to both $X$ and $Y$); see \Cref{app:confounders} for how confounders complicate simple-TR.

Notably, the risk upper-bound provided in \Cref{lem:simple-guarantees} is not tight, e.g., $\epsilon^2$ in the denominator of the transportable case can be improved through adaptive procedures. Since we are assuming that $\epsilon$ is a constant and $N \gg n$, the bounds serve the purpose for this work. Our main focus throughout is identifying which source domains contain useful information for prediction in the target, and how that information can be incorporated in learning, given the structure $\Delta$, thus we rely on the covariates overlap. On the other hand, in many theoretical work on domain adaptation, it is presumed that the sources are all useful for prediction in target, e.g., there exists a unique best hypothesis $h^\star$ for all domains \cite{ben-david-2006,mansour2009}, thus, in these work the main complexity of DA comes from lack of overlap between the covariate distributions across the domains.

\begin{wrapfigure}{R}{0.55\textwidth}
    \begin{minipage}{0.55\textwidth}
      \begin{algorithm}[H]
        \caption{simple-AD} \label{alg:AD-no-Delta}
        \begin{algorithmic}[1]
    \REQUIRE $D^{\src}, D^\star$
   \ENSURE $\mu_\AD(y\mid x) \approx P^\star(y\mid x)$
   \STATE $D^\star_{\train}, D^\star_{\test} \gets \mathrm{partition}(D^\star)$
   \STATE $\psi_{\text{share}} \gets \hat{P}(y \mid x;\, D^\star_\train \cup D^{\src})$
   \STATE $\psi_{\text{none}} \gets \hat{P}(y \mid x;\, D^\star_\train)$
    \STATE \textbf{Return} $\mu_\AD \gets \argmin_{\psi \in \{\psi_{\text{share}}, \psi_{\text{none}}\}} \ell(\psi;\, D^\star_\test)$
    \end{algorithmic}
      \end{algorithm}
    \end{minipage}
\end{wrapfigure}

We treat the simple-TR procedures (such as \Cref{alg:AD-Delta}) as the best one can do given knowledge of the structural properties of the problem. In reality, access to such structure may not be viable, and in the next example we would like to consider adaptation in situations where  $\Delta$ is unknown.

\begin{example} [Agnostic approach; $X\to Y$ case]  \label{ex:agnostic}
\normalfont
In the context of \Cref{ex:classification-no-confounding}, suppose $\Delta$ is unknown but we want guarantees close to what simple-TR achieves. Recall simple-TR pools source with target if $\Delta$ implies $P^\star(y\mid x) = P^{\src}(y\mid x)$, giving rate in $N$; otherwise rate in $n$.

Without $\Delta$, we proceed as follows. Partition $D^\star = D^\star_\train \sqcup D^\star_\test$ with $|D^\star_\train|=|D^\star_\test|=\tfrac{n}{2}$. Train two candidates: $\psi_{\text{share}}(y\mid x) := \hat{P}(y\mid x;\, D^\star_\train \cup D^{\src})$ (assumes mechanism is shared) and $\psi_{\text{none}}(y\mid x) := \hat{P}(y\mid x;\, D^\star_\train)$ (target-only). Use $D^\star_\test$ to select between them, $\mu_\AD \gets \argmin_{\psi \in \{\psi_{\text{share}}, \psi_{\text{none}}\}} \sum_{(x,y)\in D^\star_\test} \ell(\psi;y,x)$. Whichever case $\Delta$ is in, the corresponding candidate achieves simple-TR's rate; the cost of selection from a 2-element class is $\cO(\sqrt{1/n})$. \hfill $\square$
\end{example}

\Cref{ex:agnostic} shows that through a simple training-validation procedure, without access to the domain discrepancies, it is possible to achieves rates that are only slightly worse than what is achievable through explicit access to the domain discrepancies $\Delta$. We call such strategies \emph{Agnostic} throughout this work, and \Cref{alg:AD-no-Delta} summarizes this approach. What follows is a formal statement.

\begin{theorem} [simple-AD error rate] \label{thm:agnostic-AD-univariate} Let $\mu_{\TR}, \mu_\AD$ be learned using simple-TR and simple-AD (\Cref{alg:AD-Delta,alg:AD-no-Delta}), respectively. Then
\begin{equation}
    R_{P^\star}(\mu_{\AD}) = \cO\!\left(R_{P^\star}(\mu_{\TR}) + \sqrt{\tfrac{1}{n}}\right),
\end{equation}
where $n$ is the target sample size. \hfill $\square$
\end{theorem}

\section{Proofs} \label{app:proofs}

\begin{definition} [strongly convex functions] A function $f(x)$ is $m$-strongly convex if,
\begin{equation}
    f(x') \geq f(x) + \nabla f(x)^T(x'-x) + \frac{m}{2}\|x'-x\|^2, \quad \forall x,x' \in \cX,
\end{equation}
\hfill $\square$
\end{definition}

Next, we show that the risk in our problem is strongly convex under strict positivity assumption $P^\star(x,y)>\epsilon$.

\begin{lemma} [strongly convex risk.] $R_{P^\star}(\mu)$ is $\epsilon$-strongly convex w.r.t. $\mu$ under the assumption of $P^\star(x,y) > \epsilon$.
\begin{proof}
Recall the loss function $\ell(\mu;y,x) = - \log \mu(y\mid x)$. Thus, the true risk of $\mu_\TR$ can be expressed as:
\begin{align}
    R_{P^\star}(\mu_\TR) 
    &= \mathbb{E}_{P^\star}[- \log \mu_{\TR}(Y\mid X)]\\
    &= \sum_{x} P^\star(x) \cdot \sum_{y} P^\star(y\mid x) \cdot \log \frac{1}{\mu_{\TR}(y\mid x)}\\
    &= \sum_{x} P^\star(x) \cdot D_{\KL}\big(P^\star(\cdot \mid x) \| \mu_{\TR}(\cdot \mid x)\big).
\end{align}
For a fixed $x\in \cX$, the KL loss $D_{\KL}\big(P^\star(\cdot \mid x) \| \mu_{\TR}(\cdot \mid x)\big)$ is $1$-strongly convex for the interior of the simplex, i.e., under positivity. Thus, the weighted sum $\sum_{x} P^\star(x) \cdot D_{\KL}\big(P^\star(\cdot \mid x) \| \mu_{\TR}(\cdot \mid x)\big).$ with $P^\star(x)>\epsilon$ would be $\epsilon$-strongly convex \cite{boyd2004convex}. 
\end{proof}
\end{lemma}

What follows is standard high-probability bound for excess risk of ERM with strongly convex risk, adapted from \cite{kakade2009generalization}.

\begin{corollary} \label{cor:strongly-convex} Let the ERM solution be,
\begin{equation}
    \mu_{\ERM} := \hat{P}(y\mid x;D^\star) \in \argmin_{\mu:\cX \to\mathrm{simplex}^{|\cY|}} \sum_{x,y \in D^\star} - \log \mu(y\mid x),
\end{equation}
and let the true risk minimizer be,
\begin{equation} \label{eq:risk-minimizer}
    \mu_* = \argmin_{\mu:\cX \to\mathrm{simplex}^{|\cY|}} \bbE_{Y,X \sim P^\star}[- \log \mu(y\mid x)].
\end{equation}
Under $P^\star(x,y) > \epsilon$, for any $\delta>0$ the following holds with probability $1-\delta$:
\begin{equation}
    R_{P^\star}(\mu_{\ERM}) - R_{P^\star}(\mu_*) \leq \frac{8\cdot \big(\ln\frac{1}{\delta} + |\cX|\cdot |\cY| \cdot \ln(1 + \frac{n}{|\cX|\cdot |\cY|})\big)}{\epsilon\cdot n} = \cO(\frac{|\cX|\cdot |\cY|}{\epsilon \cdot n}),
\end{equation}
where $n = |D^\star|$.
\hfill $\square$
\end{corollary}

We also show the following bound for the risk of the transported estimators.

\begin{lemma} \label{lem:transported-bound} Suppose $P^\star(y\mid x) = P(y\mid x)$. Define the transported predictor as the ERM over $D \sim P(x,y)$:
\begin{equation}
    \mu_{\TR} := \hat{P}(y\mid x;D) \in \argmin_{\mu:\cX \to\mathrm{simplex}^{|\cY|}} \sum_{x,y \in D} - \log \mu(y\mid x),
\end{equation}
and let the true risk minimizer be defined in \Cref{eq:risk-minimizer}.
Suppose $P^\star(x,y),P(x,y) > \epsilon$. For any $\delta>0$ the following holds with probability $1-\delta$:
\begin{equation}
    R_{P^\star}(\mu_{\TR}) - R_{P^\star}(\mu_*) = \cO(\frac{|\cX|\cdot |\cY|}{\epsilon^2 \cdot N}),
\end{equation}
where $N = |D|$.
\begin{proof}
Equal conditionals, i.e., $P^\star(y\mid x) = P(y\mid x)$, implies that the true risk minimizer matches under both distributions, i.e.,
\begin{equation}
    \mu_* \in \argmin_{\mu} \bbE_{P^\star}[\ell(\mu;y,x)] \iff \mu_* \in \argmin_{\mu} \bbE_{P}[\ell(\mu;y,x)].
\end{equation}
Since $\mu_{\TR}$ is the solution of ERM under $P$, based on \Cref{cor:strongly-convex}, we have:
\begin{align} \label{eq:optimal-under-P}
    R_{P}(\mu_\TR) - R_{P}(\mu_*) = \cO(\frac{|\cX|\cdot |\cY|}{\epsilon \cdot N}).
\end{align}
Let $\alpha = \max_{x\in \mathcal{X}} \frac{P^\star(x)}{P(x)}$. For any $\mu:\cX \to \mathrm{simplex}^{|\cY|}$, we related the risk under $P$ and $P^\star$:
\begin{align}
R_{P^\star}(\mu_{\TR}) - R_{P^\star}(\mu_*) &= \sum_{x} P^\star(x) \cdot \sum_{y} P(y|x) \cdot (\log \mu_{*}(y|x) -\log \mu_{\TR}(y|x))\\
&= \sum_{x} \frac{P^\star(x)}{P(x)} \cdot P(x) \cdot \sum_{y} P(y|x) \cdot (\log \mu_{*}(y|x) -\log \mu_{\TR}(y|x))\\
&= \mathbb{E}_{(X,Y)\sim P}\left[\frac{P^\star(X)}{P(X)} \cdot (\log \mu_{*}(y|x) -\log \mu_{\TR}(y|x))\right]\\
& \leq \alpha \cdot \mathbb{E}_{(X,Y)\sim P}\left[\log \mu_{*}(y|x) -\log \mu_{\TR}(y|x)\right]\\
&= \alpha \cdot \big(R_{P}(\mu_{\TR}) - R_{P}(\mu_{*})\big) \\
&= \cO(\frac{\alpha \cdot |\cX|\cdot |\cY|}{\epsilon \cdot N}) = \cO(\frac{|\cX|\cdot |\cY|}{\epsilon^2 \cdot N}).
\end{align}
The last line follows from strict positivity:
\begin{equation}
    \alpha = \max_{x\in \mathcal{X}} \frac{P^\star(x)}{P(x)} \leq  \frac{\max_{x\in \mathcal{X}} P^\star(x)}{\min_{x\in \cX} P(x)} \leq \frac{1}{\epsilon}
\end{equation}
\end{proof}
\end{lemma}

\subsection{Proof of \Cref{lem:simple-guarantees}} \label{sec:proof-simple-guarantees}

If $Y \in \Delta$, the algorithm reduces to $\mu_\TR \gets \hat{P}(y\mid x;D^\star)$, i.e., ERM on target only. \Cref{cor:strongly-convex} yields the desired guarantee.

If $Y \notin \Delta$, the source $Y$ mechanism matches the target's, so $P^{\src}(y\mid x) = P^\star(y\mid x)$ and we transport the predictor from $D^{\src}$ pooled with $D^\star$. Since the pooled data has size $N+n$ with $N \gg n$, \Cref{lem:transported-bound} yields the desired result.

\subsection{Proof of \Cref{thm:agnostic-AD-univariate}} \label{sec:proof-agnostic-AD-univariate}
In \Cref{alg:AD-no-Delta} we first compute two candidate predictors corresponding to the two possible values of $\Delta_Y$: $\psi_{\text{share}} = \hat{P}(y\mid x;\, D^{\src} \cup D^\star_\train)$ and $\psi_{\text{none}} = \hat{P}(y\mid x;\, D^\star_\train)$, then use held-out target data to choose the one with smaller risk. Let
\begin{equation}
    \tilde{\mu} \in \argmin_{\mu \in \{\psi_{\text{share}},\, \psi_{\text{none}}\}} R_{P^\star}(\mu),
\end{equation}
and let $\mu_\TR$ be obtained from \Cref{alg:AD-Delta}. Consider the two cases:
\begin{enumerate}
    \item $Y \in \Delta$: $\mu_\TR$ is target-only ERM, $\mu_\TR = \hat{P}(y\mid x;D^\star)$, with rate (\Cref{cor:strongly-convex})
    \begin{equation}
        R_{P^\star}(\mu_\TR) - R_{P^\star}(\mu_*) = \cO\!\left(\tfrac{|\cX||\cY|}{\epsilon n}\right).
    \end{equation}
    The candidate $\psi_{\text{none}}$ uses $D^\star_\train$ of size $\tfrac{n}{2}$, so
    \begin{equation}
        R_{P^\star}(\tilde{\mu}) - R_{P^\star}(\mu_*) \leq R_{P^\star}(\psi_{\text{none}}) - R_{P^\star}(\mu_*) = \cO\!\left(\tfrac{|\cX||\cY|}{\epsilon n}\right).
    \end{equation}
    \item $Y \notin \Delta$: $\mu_\TR$ pools source and target, with rate (\Cref{lem:transported-bound})
    \begin{equation}
        R_{P^\star}(\mu_\TR) - R_{P^\star}(\mu_*) = \cO\!\left(\tfrac{|\cX||\cY|}{\epsilon^2 N}\right).
    \end{equation}
    The candidate $\psi_{\text{share}}$ uses the same pooled data as $\mu_\TR$ (with target half-size), giving
    \begin{equation}
        R_{P^\star}(\tilde{\mu}) - R_{P^\star}(\mu_*) \leq R_{P^\star}(\psi_{\text{share}}) - R_{P^\star}(\mu_*) = \cO\!\left(\tfrac{|\cX||\cY|}{\epsilon^2 N}\right).
    \end{equation}
\end{enumerate}
Hence
\begin{equation}\label{eq:same-rate-with-TR}
    R_{P^\star}(\tilde{\mu}) = \cO(R_{P^\star}(\mu_\TR)).
\end{equation}
Next, the empirical version $\mu_\AD$ is obtained by minimizing empirical risk over the two-element class $\{\psi_{\text{share}}, \psi_{\text{none}}\}$ using $D^\star_\test$ of size $\tfrac{n}{2}$. Uniform convergence over a finite hypothesis class \citep{vapnik1971uniform,shalev2014understanding} gives, with probability $1-\delta$,
\begin{equation}
    R_{P^\star}(\mu_\AD) - R_{P^\star}(\tilde{\mu}) \leq \sqrt{\tfrac{\log 2}{n}} + \sqrt{\tfrac{\log(1/\delta)}{n}} = \cO\!\left(\sqrt{\tfrac{1}{n}}\right).
\end{equation}
Combined with \Cref{eq:same-rate-with-TR},
\begin{equation}
    R_{P^\star}(\mu_\AD) = \cO\!\left(R_{P^\star}(\mu_\TR) + \sqrt{\tfrac{1}{n}}\right).
\end{equation}

\subsection{Proof of \Cref{lem:structure-informed-multi}}

The proof follows the logic of the proof of \Cref{lem:simple-guarantees} (\Cref{sec:proof-simple-guarantees}). In the transportable case ($Y \notin \Delta$) the data used for estimation of $\mu_{\TR}$ is pooled from $D^{\src}$ of size $N$ and $D^\star$ of size $n$, with $N \gg n$. The conditional distribution to be estimated is $\mu_{\TR}(y\mid \pa^\star_Y)$, and for $c = |\Pa^\star_Y|$, \Cref{lem:transported-bound} gives
\begin{equation}
    R_{P^\star}(\mu_{\TR}) - R_{P^\star}(\mu_{*}) = \cO\!\left(\tfrac{|\cX|^c |\cY|}{\epsilon^2 N}\right).
\end{equation}
In the non-transportable case ($Y \in \Delta$), $\mu_{\TR}$ uses only target data, so by \Cref{cor:strongly-convex}
\begin{equation}
    R_{P^\star}(\mu_{\TR}) - R_{P^\star}(\mu_{*}) = \cO\!\left(\tfrac{|\cX|^c |\cY|}{\epsilon n}\right).
\end{equation}

\subsection{Proof of \Cref{thm:structure-informed-sequence}} \label{sec:proof-structure-informed-sequence}

The query of interest $P^\star(v_{T_*}\mid v_{1:M})$ can be computed through the following formula by introducing the intermediate variables $V_{M+1:T_*-1}$ and then marginalizing them out:
\begin{equation}
    P^\star(v_{T_*}\mid v_{1:M}) = \sum_{v_{M+1:T_*-1}} P^\star(v_{M+1:T_*} \mid v_{1:M}).
\end{equation}
Following the causal order, and the causal diagram of the target domain, we can write $P^\star(v_{M+1:T_*} \mid v_{1:M})$ as a product of conditionals on the parents:
\begin{equation}
    P^\star(v_{T_*}\mid v_{1:M}) = \sum_{v_{M+1:T_*-1}} \prod_{i=M+1}^{T_*} P^\star(v_i \mid \pa^\star_i).
\end{equation}
To transport the above, we attempt a multi-cause problem instance at every target position $i$: if there exists a source position $i' \in [T]$ such that $\Delta(i, i') = 0$,
\begin{equation}
    P^\star(V^\star_i = y \mid \Pa^\star_i = \bx) = P^{\src}(V^{\src}_{i'} = y \mid \Pa^{\src}_{i'} = \bx).
\end{equation}
We pool data across all source positions $i'$ with $\Delta(i, i') = 0$ to estimate $P^\star(V^\star_i = y \mid \Pa^\star_i = \bx)$. In \Cref{alg:AD-Delta-sequence}, $\cJ_i \subseteq [T]$ denotes this subset of source positions. The pooled data $D^{\TR}_i$ has effective size $|\cJ_i| \cdot N + n$, where the $n$ term is contributed by the target. ERM on $D^{\TR}_i$ gives $\mu_\TR^i := \hat{P}(y\mid \bx; D^{\TR}_i)$. Composing these predictors and marginalizing the intermediate variables yields an estimator for $P^\star(v^\star_{T_*}\mid v^\star_{1:M})$:
\begin{equation}
    \mu_{\TR}(v_{T_*} \mid v_{1:M}) = \sum_{v_{M+1:T_*-1}} \prod_{i=M+1}^{T_*} \mu_{\TR}^i(Y = v_{i} \mid \bX = \pa^\star_i).
\end{equation}
Next, we decompose the risk of $\mu_{\TR}$ in terms of the risk of the predictors different positions:
\begin{align}
    R_{P^\star}(\mu_{\TR})
    &= \bbE_{P^\star}[-\log \mu_{\TR}(v_{T_*}\mid v_{1:M})] && (\ell(\mu;y,\bx) = -\log \mu(y\mid \bx))\\
    &= \bbE_{P^\star}\big[ \bbE_{P^\star}[-\log \mu_{\TR}(v_{T_*}\mid v_{1:M})\mid v_{M+1:T_*-1}] \big] && (\text{law of iterated expectation.})\\
    &= \bbE_{P^\star}\big[ \bbE_{P^\star}[-\log \sum_{v_{M+1:T_*-1}} \prod_{i=M+1}^{T_*} \mu^i_{\TR}(v_i\mid \pa^\star_i)\mid v_{M+1:T_*-1}] \big] && (\text{intermediate variables.})\\
    &\leq \bbE_{P^\star}\big[ \bbE_{P^\star}[\sum_{{v_{M+1:T_*-1}}} -\log \prod_{i=M+1}^{T_*} \mu^i_{\TR}(v_i\mid \pa^\star_i)\mid v_{M+1:T_*-1} ]\big] && (\text{concavity of $\log$ \& Jensen ineq.})\\
    &= \bbE_{P^\star}\big[ \sum_{{v_{M+1:T_*-1}}}\sum_{i=M+1}^{T_*} \bbE_{P^\star}[- \log \mu^i_{\TR}(v_i\mid \pa^\star_i)\mid v_{M+1:T_*-1} ]\big] && (\text{linearity of expectation})\\
    &= \sum_{i=M+1}^{T_*} \bbE_{P^\star}\big[ \sum_{{v_{M+1:i-1}}} \bbE_{P^\star}[- \log \mu^i_{\TR}(v_i\mid \pa^\star_i)\mid v_{M+1:i} ]\big] && (\text{Markovianity})\\
    &= \sum_{i=M+1}^{T_*} \bbE_{P^\star}[- \log \mu^i_{\TR}(v_i\mid \pa^\star_i)] && (\text{marginalize interm. vars.})\\
    &= \sum_{i=M+1}^{T_*} R_{P^\star}(\mu^i_{\TR}) && (\text{risk of sub-task})
\end{align}
Let $\mu^i_* := P^\star(v_i\mid \pa^\star_i)$, so that,
\begin{equation}
    \mu_*(v_{T_*}\mid v_{1:M}) = \sum_{v_{M+1:T_*-1}} \prod_{i=M+1}^{T_*} \mu^i_*(v_i \mid \pa^\star_i).
\end{equation}
Due to \Cref{lem:transported-bound} and \Cref{cor:strongly-convex}, we have the following risk bound for the predictor at position $i$:
\begin{equation}
    R_{P^\star}(\mu_{\TR}^i) - R_{P^\star}(\mu_*^i) \leq \cO(\frac{|\cV|^{|\Pa^\star_i| + 1}}{|\cJ_i|\cdot \epsilon^2 \cdot N + \epsilon \cdot n}).
\end{equation}
Therefore, we have the bound,
\begin{align} \label{eq:bound-sequence-risk}
    R_{P^\star}(\mu_{\TR}) - R_{P^\star}(\mu_*)
    &\leq \sum_{i=M+1}^{T_*} R_{P^\star}(\mu^i_{\TR}) - R_{P^\star}(\mu^i_*)\\
    &=  \cO\big(\sum_{i=M+1}^{T_*} \frac{|\cV|^{|\Pa^\star_i| + 1}}{|\cJ_i|\cdot \epsilon^2 \cdot N + \epsilon \cdot n} \big).
\end{align}
Let,
\begin{equation}
    \cI = \{i\in \{M+1,\dots,T_*\} \text{ s.t. }\cJ_{i} \neq \emptyset\},
\end{equation}
denote the positions for which $P^\star(v_i\mid \pa^\star_i)$ is transportable. Also, let $c=\max_{i\in[T_*]}|\Pa^\star_i|$. We have,
\begin{equation} \label{eq:refined-rate-sequence}
    R_{P^\star}(\mu_{\TR}) - R_{P^\star}(\mu_*) = \cO\big( \frac{|\cI|\cdot|\cV|^{c + 1}}{\epsilon^2\cdot N} + \frac{(T_*-M-|\cI|)\cdot|\cV|^{c + 1}}{\epsilon\cdot n} \big).
\end{equation}
The latter justifies the claim of \Cref{thm:structure-informed-sequence}. We discuss the different rates achievable above in \Cref{app:structure-informed-rates}.

\subsection{Proof of \Cref{thm:agnostic-sequential}}

The proof follows the logic of the proof of \Cref{thm:agnostic-AD-univariate} (\Cref{sec:proof-agnostic-AD-univariate}). \Cref{alg:agnostic-sequence} iterates over partitions of $\cE = [T] \cup \{1^\star,\dots,T_*^\star\}$ into clusters; each cluster identifies a set of source and target positions assumed to share a mechanism. For each partition (encoding a candidate $\Delta$) and each pair of causal diagrams $(\cG^{\src}, \cG^\star)$, we run circuit-TR (\Cref{alg:AD-Delta-sequence}) to obtain a candidate estimator for $P^\star(v^\star_{T_*}\mid v^\star_{1:M})$ added to $\cH$.

For every structurally consistent $(\Delta, \cG^{\src}, \cG^\star)$ there is a candidate in $\cH$ matching the data circuit-TR would use. Hence
\begin{equation}
    R_{P^\star}(\tilde{\mu}) = \cO(R_{P^\star}(\mu_{\TR})), \quad \tilde{\mu} \in \argmin_{\mu \in \cH} R_{P^\star}(\mu).
\end{equation}
We obtain $\mu_\AD$ as the empirical risk minimizer in $\cH$ on held-out target data $D^\star_\test$ of size $\tfrac{n}{2}$:
\begin{equation}
    \mu_{\AD} \in \argmin_{\mu \in \cH} \frac{1}{|D^\star_\test|} \sum_{(x,y)\in D^\star_\test} \ell(\mu;y,x),
\end{equation}
giving (by uniform convergence)
\begin{equation}
    R_{P^\star}(\mu_{\AD}) - R_{P^\star}(\tilde{\mu}) = \cO\!\left(\sqrt{\tfrac{\log|\cH|}{n}}\right).
\end{equation}
Bounding the size of $\cH$:
\begin{equation}
    |\cH| \leq \underbrace{(T+T_*)^{T+T_*}}_{\text{partitions of $\cE$}} \cdot \underbrace{((2T)!)^{T} \cdot ((2T_*)!)^{T_*}}_{\text{causal diagrams $\cG^{\src}, \cG^\star$}}
\end{equation}
which yields, taking $T$ and $T_*$ of the same order (denoted $T_*$ for the bound),
\begin{equation}
    \log|\cH| = \cO(T_*^3 \log T_*).
\end{equation}
This gives the claim of \Cref{thm:agnostic-sequential}:
\begin{equation}
    R_{P^\star}(\mu_\AD) = \cO\!\left(R_{P^\star}(\mu_\TR) + \sqrt{\tfrac{T_*^3 \log T_*}{n}}\right).
\end{equation}

\subsection{Proof of \Cref{cor:fast-slow-threshold}} \label{sec:proof-fast-slow}

By \Cref{thm:agnostic-sequential}, $R_{P^\star}(\mu_\AD) = \cO\!\big(R_{P^\star}(\mu_\TR) + \sqrt{T_*^3 \log T_* / n}\big)$. Take $T_* = L$, the size of the smallest target circuit composable from source-learnable modules computing $P^\star(y\mid \bx)$. In this regime $R_{P^\star}(\mu_\TR)$ decays in $N$ alone (\Cref{thm:structure-informed-sequence} with $|\cI| = T_*-M$), so for $N \gg n$ the dominant term is $\sqrt{L^3 \log L / n}$. This vanishes as $n\to\infty$ iff $L^3 \log L = o(n)$, i.e., $L = o\!\big((n/\log n)^{1/3}\big) = \cO(\sqrt[3]{n})$ up to polylog factors. When $L$ is constant, $\sqrt{L^3\log L/n} = \cO(1/\sqrt{n})$, recovering the classical fast rate.

\subsection{Proof of \Cref{thm:pretraining}}

Define
\begin{equation}
    \mu_{\theta}(v_i\mid v_{1:i-1}) := \prob_\theta\!\big(v_i \mid {A_\theta^{\src}}_{i,\cdot}\cdot v_{1:T}; \Phi_\theta(i)\big).
\end{equation}
We can rewrite the objective of \Cref{eq:pretraining} as
\begin{equation}
    \cL(\theta) := \lambda\cdot \underbrace{\Big(d_{\theta} + \sum_{i,i' = 1}^T {A^{\src}_\theta}_{i,i'}\Big)}_{\text{penalty}} + \underbrace{\sum_{i=1}^T \bbE_{P^{\src}}\!\big[ D_{\KL}\big(P^{\src}(\cdot\mid V_{1:i-1}) \,\|\, \mu_{\theta}(\cdot\mid V_{1:i-1})\big) \big]}_{\text{match with source distribution}}.
\end{equation}
In words, the objective ensures that the solution matches the source at all conditionals while preferring parameters with smaller mechanism-indicator range $d$ and fewer edges in $A^{\src}$.

For any $\lambda > 0$ small enough that the data term strictly dominates the penalty among optimal parameterizations (i.e., $\lambda$ is below the gap between the best feasible KL and the second-best on a different parent-set realization), minimizing $\cL(\theta)$ forces $A^{\src}$ to have a one entry at the position of the true parents in each row (since $P^{\src}(\cdot\mid v_{1:i-1}) = P^{\src}(\cdot\mid \pa^{\src}_i)$); the $\lambda\|A^{\src}\|_1$ penalty keeps no additional one entries, satisfying \Cref{eq:condition-graphs}.

If two source positions $i, i'$ share a mechanism, $P^{\src}(v_i\mid \pa^{\src}_i) = P^{\src}(v_{i'}\mid \pa^{\src}_{i'})$. To satisfy \Cref{eq:condition-mechanisms}, $\Phi: [T]\to[d]$ must map $i$ and $i'$ to the same value, so that $\mu_\theta(v_i\mid\pa^{\src}_i) = \mu_\theta(v_{i'}\mid\pa^{\src}_{i'})$. Minimizing $\lambda\cdot d$ enforces this. Finally, with \Cref{eq:condition-mechanisms,eq:condition-graphs} satisfied, minimizing the divergence forces \Cref{eq:condition-prob}.

\subsection{Proof of \Cref{thm:fine-tuning-rate}}

The parameters learned in the fine-tuning stage are:
\begin{enumerate}
    \item The target mechanism indicator $\Phi^\star: [T_*] \to [d_{\theta^\src}]$.
    \item The target parent matrix $A^\star \in [0,1]^{T_* \times T_*}$.
    \item The target-only predictors $\mu^\star_i(v_i\mid v_{1:i-1})$.
    \item The transport indicators $s_1,\dots,s_{T_*} \in [0,1]$.
\end{enumerate}
Once the pretrained parameters $\theta^\src$ satisfy \Cref{eq:condition-mechanisms,eq:condition-graphs,eq:condition-prob}, consider the following parameter values for fine-tuning. Let $A^\star$ encode the true target causal diagram $\cG^\star$. For each target position $i$ where the conditional $P^\star(v_i\mid\pa^\star_i)$ is transportable (i.e., $\Delta(i, i') = 0$ for some source position $i'$), set $s_i = 1$ and $\Phi^\star(i) = \Phi(i')$. For non-transportable positions, set $s_i = 0$ to use $\mu^\star_i(v_i\mid v_{1:i-1})$ trained on $D^\star_\train$ of size $\propto n$. Let $\tilde{\theta}$ denote this assignment. We have,
\begin{equation}
    R_{P^\star}\!\big(\mu^{\tilde{\theta}}_\ft(v_{T_*}\mid v_{1:M})\big)  = \cO\!\big(R_{P^\star}(\mu_\TR(v_{T_*}\mid v_{1:M}))\big),
\end{equation}
Since this set of values for the parameters corresponds to the circuit-TR solution. We discretize the fine-tuning parameter space into a set $\cH$ of points , and consider only binary parent matrices and binary transport indicators. We can ensure that $\tilde{\theta}$ lies on this grid, among
\begin{equation}
    |\cH| \leq \underbrace{T_*^{T_*}}_{\text{parent matrix}} \cdot \underbrace{d^{T_*}}_{\text{target mech. ind.}} \cdot \underbrace{2^{T_*}}_{\text{TR indicator}}
\end{equation}
We use the held-out target data to obtain the best of these candidates,
\begin{equation}
    \theta^\star \in \argmin_{\theta \in \cH} \frac{1}{|D^\star_\test|}\cdot \sum_{v\in D^\star_\test} -\log \mu^\theta_\ft(v_{T_*}\mid v_{1:M})
\end{equation}
Compared to the best in class parameter set $\tilde{\theta}$, we would have an excess risk bounded as,
\begin{equation}
    R_{P^\star}(\mu_{\ft}^{\theta^\star}) - R_{P^\star}(\mu_{\ft}^{\tilde{\theta}}) = \cO\!\left(\sqrt{\tfrac{\log |\cH|}{n}}\right) = \cO\!\left(\sqrt{\tfrac{T_* \log T_*}{n}}\right).
\end{equation}
This proves,
\begin{align}
    R_{P^\star}(\mu_{\ft}^{\theta^\star})
    &= \cO\!\left(R_{P^\star}(\mu_\TR) + \sqrt{\tfrac{T_* \log T_*}{n}}\right) \\
    &= \cO\!\left(R_{P^\star}(\mu_\TR) + \sqrt{\tfrac{T_*^3 \log T_*}{n}}\right) = \cO(R_{P^\star}(\mu_{\AD}))
\end{align}
\section{Challenges due to unobserved confounders} \label{app:confounders}

In this section we discuss how presence of confounders raises challenges in structure-informed and agnostic DA. We show that even in a very simple case, despite having the same causal functions generating $Y$ from $X$ and the unobserved variables between a source and target, transport is impossible. 

\begin{figure}[h]
    \centering
  \includegraphics[scale=0.5]{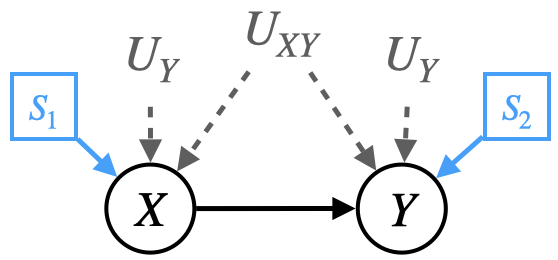}
  \caption{Selection diagram $\cG^{\Delta}$}
  \label{fig:bow}
\end{figure}

Let $X$ be a single binary covariate, and $Y$ be a binary label. Consider two source domains defined by the following SCMs:
\begin{align*}
    &\cM^1 : \begin{cases}
        P^1(\bU): \begin{cases}
            U_X \sim
                \mathrm{Bern}(0.2)  \\
            U_Y \sim \mathrm{Bern}(0.05)\\
            U_{XY} \sim \mathrm{Bern}(0.95) \\
        \end{cases}\\
    \cF^1: \begin{cases}
    X \gets U_X \oplus U_{XY}\\
    Y \gets 
        (X \oplus U_{XY}) \oplus U_Y\\
    \end{cases}
    \end{cases} 
    &\cM^2 : \begin{cases}
        P^2(\bU): \begin{cases}
            U_X \sim \mathrm{Bern}(0.9) \\
            U_Y \sim \mathrm{Bern}(0.05) \\
            U_{XY} \sim \mathrm{Bern}(0.95) 
        \end{cases}\\
    \cF^2: \begin{cases}
    X \gets U_X \oplus U_{XY} \\
    Y \gets
        (X \oplus U_{XY}) \vee U_Y \\
    \end{cases}
    \end{cases} 
\end{align*}

Suppose the target domain is represented by the following SCM:

\begin{align*}
    \cM^\star : \begin{cases}
        P^\star(\bU): \begin{cases}
            U_X \sim \mathrm{Bern}(0.9)\\
            U_Y \sim \mathrm{Bern}(0.05)\\
            U_{XY} \sim \mathrm{Bern}(0.95)
        \end{cases}\\
    \cF^\star: \begin{cases}
    X \gets U_X \oplus U_{XY}\\
    Y \gets (X \oplus U_{XY}) \oplus U_Y
    \end{cases}
    \end{cases}
\end{align*}

This setup induces the domain discrepancy sets $\Delta_{1,*} = \{X\}, \Delta_{2,*} = \{Y\}$; the corresponding selection diagram $\cG^\Delta$ is shown in \Cref{fig:bow}. Although the mechanism of $Y$ is invariant between $\pi^\star,\pi^1$, we can not transport $P^\star(y\mid x)$; this follows from completeness results in transportability \citep{pearl2011transportability,correa2020gtr}, that there exist source and target SCMs compatible with $\cG^\Delta$, however, we would have $P^\star(y\mid x) \neq P^1(y\mid x)$, e.g., $P^\star(Y=1\mid x=1) \approx 0.34$ but $P^1(Y=1\mid x=1) \approx 0.06$.

Nonetheless, one can derive a tight bound for $P^\star(y\mid x)$ \citep{balke1997bounds}; in particular, we can compute $l_{y\mid x}, u_{y\mid x}$ such that,
\begin{equation}
    P^\star(y\mid x) \in [l_{y\mid x}, u_{y\mid x}].
\end{equation}
Moreover, this bound is tight, in the sense that for every $q \in [l_{y\mid x}, u_{y\mid x}]$, there exist source and target SCMs that entail $P^1,P^2$ over $X,Y$, induce the selection diagram $\cG^\Delta$, such that we have $P^\star(y\mid x) = q$. Deriving these bounds is called partial-transportability, and is studied in \citep{jalaldoust2024partial}. Even without any target data, although $P^\star(y\mid x)$ is not transportable, we can compute a subset of conditional distributions that contains it through partial-transportability methods. In particular, let
\begin{equation}
    \cP^\star = \{P^{\cM^\star_0}(x,y) \text{ s.t. } \cM^1_0, \cM^2_0, \cM^\star_0 \text{ entail } P^1(x,y),P^2(x,y), \text{ and induce }\cG^{\Delta}\}.
\end{equation}
By definition, $P^\star(x,y) \in \cP^\star(y\mid x)$ holds under the structural assumptions encoded in $\cG^\Delta$. Thus, even without any target data, we can achieve non-trivial risk by solving the following min-max problem:
\begin{equation}
    \mu_{\pTR} := \argmin_{\mu:\cX \to \mathrm{simplex}^{|\cY|}} \max_{\tilde{P} \in \cP^\star } R_{\tilde{P}}(\mu).
\end{equation}
A solution to this problem is called causal robust optimization (CRO) by \citet{jalaldoust2024partial}. Notably, with access to target data we can not significantly improve the risk achieved by $\mu_{\pTR}$; we elaborate on this point below.

Since $X,Y$ are binary, distributions over them can be parameterized by a point in $\mathrm{simplex}^3$. Consider $\cP^\star$ as a subset of the simplex, and suppose $P^\star(x,y)$ lies in the \emph{interior} of $\cP^\star$, that is, there exists $\delta >0 $ such that a $\delta$-ball around $P^\star(x,y)$ is contained inside $\cP^\star$. Using vanilla ERM on the target data we can estimate the joint distribution $P^\star(x,y)$, and form a confidence set that contains it. The diameter of this confidence set shrinks to zero by increasing the target data, due to uniform convergence properties of the ERM estimator. Take the target sample size $n_\delta$ such that the ERM confidence set containing $P^\star(x,y)$ has a diameter smaller than $\delta$, and therefore, is contained entirely inside $\cP^\star$. Note that source data together with the structure $\cG^{\Delta}$ only implies $P^\star(x,y) \in \cP^\star$, so for target data of size $n>n_{\delta}$, the information $P^\star(x,y) \in \cP^\star$ would be obsolete, since ERM on target data only achieves a strictly better confidence set for $P^\star(x,y)$. 

The above is unlike the unconfounded scenario; in the transportable case, the structure-informed procedure yields zero-shot generalization (i.e., rates in terms of the source data) that cannot be beaten by vanilla ERM on the target data for any amount of target data.

\begin{figure}[t]
    \centering
  \includegraphics[scale=0.35]{figs/ERMvsCRO.png}
  \caption{Risk of ERM on target data alone compared to ERM constrained to the partially transported set $\mathcal{P}^\star$ constructed from $P^1(x,y),P^2(x,y),\cG^{\Delta}$. As shown here, for smaller target data size there is a meaningful gap between the risk of the two estimators, but for larger $n$ this gap closes.}
  \label{fig:ERMvsCRO}
\end{figure}

To summarize, presence of confounders creates situations where certain quantities are partially transportable. This, in turn, enables non-trivial but imperfect generalization without any target data, and such situations can not occur without confounders. The immediate advantage of source data through partial transportability vanishes as the size of the target data grows, and this is shown in \Cref{fig:ERMvsCRO}.
\hfill $\square$

\section{More details on structure-informed rates} \label{app:structure-informed-rates}

\begin{figure}[t]
    \centering
    \begin{tikzpicture}[scale=1.2]
        \draw[-{Latex[scale=1.2]}] (-1,0) -- (10,0) node[below right] {Risk};
        
        \draw (0,0.2) -- (0,-0.2) node[above=0.8cm] {All-TR} node[below=0.2cm] {$\frac{(T-M) \cdot |V|^{C+1}}{\varepsilon^2 N}$};
        
        \draw (3,0.2) -- (3,-0.2) node[above=0.8cm] {$(T-1)$-TR} 
        node[below=0.2cm] {$\frac{(T - M - 1) \cdot |V|^{C+1}}{\varepsilon^2 N} + \frac{|V|^{C+1}}{\varepsilon \cdot n}$};
        
        \draw (5,0.2) -- (5,-0.2) node[above=0.8cm] {$\ldots$} node[below=0.2cm] {$\ldots$};
        
        \draw (7,0.2) -- (7,-0.2) node[above=0.8cm] {$1$-TR} node[below=0.2cm] {$\frac{|V|^{C+1}}{\varepsilon^2 N} + \frac{(T-M-1)\cdot |V|^{C+1}}{\varepsilon \cdot n}$};
        
        \draw (9.5,0.2) -- (9.5,-0.2) node[above=0.8cm] {None-TR} 
        node[below=0.2cm] {$\frac{T \cdot |V|^{C+1}}{\varepsilon \cdot n}$};
    \end{tikzpicture}
    \caption{A schematic of risks obtained via the structure-informed procedure (\Cref{alg:AD-Delta-sequence}. In cases where all conditionals $P^\star(v_i\mid v_{1:i-1})$ are transportable, we obtain a rate proportionate to $\frac{T-M}{\epsilon \cdot N}$. As more and more conditionals are non-transportable, they need to be estimated from the target data, adding a cost proportionate to $\frac{1}{n}$ for every non-transportable term. In the extreme case that no conditional is transportable, i.e., all target mechanisms are novel, we incur a risk proportionate to $\frac{T}{n}$.}
    \label{fig:rates-spectrum}
\end{figure}

In this section, we expand upon the possible rates in sequence adaptation via the structure-informed procedure. Please view the proof of \Cref{thm:structure-informed-sequence} (\Cref{sec:proof-structure-informed-sequence}). We reduce the problem of estimating $P^\star(v_T\mid v_{1:M})$ to estimating,
\begin{equation}
    P^\star(v_{M+1:T} \mid v_{1:M}) = \prod_{i=M+1}^T P^\star(v_i\mid v_{1:i-1}).
\end{equation}
Each conditional is either transported from a source position (if there exists $i'$ such that $\Delta(i, i') = 0$) or estimated from target data alone. In the former case, the excess risk for $P^\star(v^\star_i\mid \pa^\star_i)$ is $\tfrac{|\cV|^{c+1}}{\epsilon^2 N}$ (desirable since $N$ is large); in the latter, $\tfrac{|\cV|^{c+1}}{\epsilon n}$. The joint risk depends on how many of the $T_*-M$ target conditionals are transported, giving the spectrum of rates in \Cref{fig:rates-spectrum}. When $N \gg n$, fast rates require all conditionals transported, with progressively slower rates per non-transportable component. The figure also bounds structure-agnostic adaptation: by \Cref{thm:agnostic-sequential}, agnostic risk is within $\sqrt{T_*^3 \log T_* / n}$ of these rates, vocabulary-independent.

\section{Simulation setup and reproducibility} \label{app:experiments}

The controlled simulation \Cref{sec:optimization} involves a source domain and a target domain with sequences of length 10. To handle more than one parent, we follow the design discussed in \Cref{app:multiple-parents-sequence}. We run simulation in a setting where each variable has at most two parents randomly selected from the previous variables in the causal order. Note that the causal diagrams of the source and target do not match necessarily, and are determined randomly and independently. The causal modules that determine the value of the variables are also drawn at random, from a pool containing null-ary, unary and binary noisy operators;
\begin{equation}
    \cF = \{g_{\mathrm{unif}}, g_{\mathrm{copy}}, g_{+1}, g_{-1}, g_{\times 2}, g_{\mathrm{sum}}, g_{\mathrm{min}}, g_{\mathrm{max}}, g_{\mathrm{subtract}}, g_{\mathrm{mult}}\}.
\end{equation}

\begin{figure}[h] 
  \centering
  \begin{tikzpicture}[
    vertex/.style={circle, draw, minimum size=0.6cm, inner sep=1pt, font=\scriptsize},
    mechanism/.style={font=\tiny, below=2pt},
    edge label/.style={font=\scriptsize}
  ]

  \foreach \i in {1,...,10} {
    \node[vertex] (V\i) at ({(\i-1)*1.4}, 0) {$V_{\i}$};
  }

  \node[mechanism] at (V1.south) {$\mathrm{unif}$};
  \node[mechanism] at (V2.south) {$\times 2$};
  \node[mechanism] at (V3.south) {$\mathrm{min}$};
  \node[mechanism] at (V4.south) {$\mathrm{sum}$};
  \node[mechanism] at (V5.south) {$\mathrm{min}$};
  \node[mechanism] at (V6.south) {$\mathrm{subtract}$};
  \node[mechanism] at (V7.south) {$\mathrm{min}$};
  \node[mechanism] at (V8.south) {$\mathrm{sum}$};
  \node[mechanism] at (V9.south) {$\mathrm{min}$};
  \node[mechanism] at (V10.south) {$\mathrm{sum}$};

  \draw[->, thick] (V1) to[bend left=0]  (V2);
  \draw[->, thick] (V1) to[bend left=30]  (V3);
  \draw[->, thick] (V2) to[bend right=0] (V3);

  \draw[->, thick] (V1) to[bend left=30] (V4);
  \draw[->, thick] (V2) to[bend left=30] (V4);

  \draw[->, thick] (V1) to[bend left=50] (V5);
  \draw[->, thick] (V4) to[bend right=0]  (V5);

  \draw[->, thick] (V1) to[bend left=30] (V6);
  \draw[->, thick] (V5) to[bend left=0]  (V6);

  \draw[->, thick] (V1) to[bend left=30] (V7);
  \draw[->, thick] (V5) to[bend left=30]  (V7);

  \draw[->, thick] (V4) to[bend left=30] (V8);
  \draw[->, thick] (V5) to[bend left=30]  (V8);

  \draw[->, thick] (V4) to[bend left=30] (V9);
  \draw[->, thick] (V6) to[bend left=30]  (V9);

  \draw[->, thick] (V3) to[bend left=30] (V10);
  \draw[->, thick] (V4) to[bend left=40] (V10);

  \end{tikzpicture}
  \caption{Causal diagram and operators corresponding to the source domain.} 
  \label{fig:graph-experiment-T=10-source}
\end{figure}

\begin{figure}[h] 
  \centering
  \begin{tikzpicture}[
    vertex/.style={circle, draw, minimum size=0.6cm, inner sep=1pt, font=\scriptsize},
    mechanism/.style={font=\tiny, below=2pt},
    edge label/.style={font=\scriptsize}
  ]

  \foreach \i in {1,...,10} {
    \node[vertex] (V\i) at ({(\i-1)*1.4}, 0) {$V_{\i}$};
  }

  \node[mechanism] at (V1.south) {$\mathrm{unif}$};
  \node[mechanism] at (V2.south) {$\times 2$};
  \node[mechanism] at (V3.south) {$\mathrm{subtract}$};
  \node[mechanism] at (V4.south) {$\mathrm{sum}$};
  \node[mechanism] at (V5.south) {$\mathrm{subtract}$};
  \node[mechanism] at (V6.south) {$\mathrm{sum}$};
  \node[mechanism] at (V7.south) {$\mathrm{subtract}$};
  \node[mechanism] at (V8.south) {$\mathrm{sum}$};
  \node[mechanism] at (V9.south) {$\mathrm{sum}$};
  \node[mechanism] at (V10.south) {$\mathrm{min}$};

  \draw[->, thick] (V1) to[bend left=0]  (V2);
  \draw[->, thick] (V1) to[bend left=30]  (V3);
  \draw[->, thick] (V2) to[bend right=0] (V3);

  \draw[->, thick] (V2) to[bend left=30] (V4);
  \draw[->, thick] (V3) to[bend right=0] (V4);

  \draw[->, thick] (V1) to[bend left=50] (V5);
  \draw[->, thick] (V4) to[bend right=0]  (V5);

  \draw[->, thick] (V1) to[bend left=30] (V6);
  \draw[->, thick] (V5) to[bend left=0]  (V6);

  \draw[->, thick] (V4) to[bend left=30] (V7);
  \draw[->, thick] (V5) to[bend left=30]  (V7);

  \draw[->, thick] (V2) to[bend left=30] (V8);
  \draw[->, thick] (V6) to[bend left=30]  (V8);

  \draw[->, thick] (V4) to[bend left=30] (V9);
  \draw[->, thick] (V7) to[bend left=30]  (V9);

  \draw[->, thick] (V7) to[bend left=30] (V10);
  \draw[->, thick] (V9) to[bend left=0] (V10);

  \end{tikzpicture}
  \caption{causal diagram and operators corresponding to the target domain.} \label{fig:graph-experiment-T=10-target}
  \label{fig:causal-row-readable}
\end{figure}

The structure we investigated is shown in \Cref{fig:graph-experiment-T=10-source,fig:graph-experiment-T=10-target}, specifying the target and source SCMs, respectively. Next, we investigate pretraining and fine-tuning in these context of this SCMs.

\subsection{Reproducibility} \label{sec:reproducibility}

\paragraph{Data generation.} Each pair is constructed by sampling a random source SCM with $T$ positions; the first position has the $\mathrm{unif}$ operator (no parents), each subsequent position has up to two parents drawn uniformly from earlier positions, and the operator is drawn uniformly from arity-matched candidates in the pool $\cF$ (\Cref{fig:graph-experiment-T=10-source}). Operators apply with probability $1-p_\mathrm{noise}$ and emit a uniform-random value with probability $p_\mathrm{noise}=0.1$. The transportable target SCM is constructed identically, restricted to operators that appear in the realised source pool (so every target mechanism has at least one source position to transport from). The slow-adaptation variant of \Cref{fig:adapt-T=10}(c) is the same construction except the last target operator is replaced by binary multiplication mod $|\cV|$, which is held outside $\cF$, so no source position transports to it.

\paragraph{Symbolic implementation.} Algorithms~\ref{alg:AD-Delta-sequence}--\ref{alg:agnostic-sequence} are realised directly: per-position conditionals are estimated as Laplace-smoothed (additive constant $0.5$) histograms from pooled data; the chain-rule query in \Cref{eq:adapt-final} is evaluated via Monte-Carlo forward sampling with $200$ rollouts per held-out evaluation row. Circuit-AD's enumeration over partitions of $[T]\cup[T_*]$ is exact for the small $T_*$ used in our sweeps.

\paragraph{Sweeps and seeds.} Source size $N\!=\!5\!\times\!10^4$ throughout. Seeds: experiment A2 uses 10 seeds for the symbolic curves; experiments B, C, D, and the gradient-based relaxation in A2 use 3-5 seeds. The held-out validation set for query NLL has $1000$ rows. Within each cell of size $n$, target data is split $70$/$30$ between candidate fitting and held-out selection, with at least one selection sample even at very small $n$.

\paragraph{Gradient-based relaxation.} The architecture of \Cref{sec:pre-training,sec:optimization} is implemented in PyTorch with hidden width $r=64$ and cluster-codomain size $d=12$. The mechanism indicator $\Phi$ is a single linear head over position embeddings followed by softmax with temperature $0.5$; the parent selector uses two causal-masked attention heads ($r$-dim queries/keys, softmax temperature $0.5$); the universal predictor is a two-layer MLP from concatenated soft-parent values and cluster soft assignment to $|\cV|$ logits. Pretraining minimises per-position next-token cross-entropy on the source sequences with weight decay $0$, AdamW lr $10^{-3}$, batch size $256$, $200$ epochs, plus penalties $\lambda_{A^{\src}}\!\cdot\!\|A^{\src}\|_1$ and $\lambda_{\Phi}\!\cdot\!H(\Phi)$ both set to $10^{-3}$. Fine-tuning unfreezes only the target-side position embeddings, $\Phi^\star$ head, and target query/key matrices, trains for $200$ epochs (with the previous $n$'s state warm-starting subsequent steps inside a $(\text{condition}, \text{seed})$ pair), AdamW lr $5\!\times\!10^{-4}$. Transport indicators are then optimised on the held-out slice for $300$ steps with Adam lr $5\!\times\!10^{-2}$. Source-side state is cached per $(T_{\src}, |\cV|, \mathrm{seed})$ and reused across the three conditions of \Cref{fig:adapt-T=10}. All hyperparameters were chosen as conventional defaults for small attention-style architectures and were fixed across all conditions and figures; we did not perform a per-condition or per-figure hyperparameter search.

\paragraph{Compute.} All experiments reported here run on CPU (one M-class workstation core); GPU is not required. Symbolic experiment A2 with $10$ seeds completes in $\approx\!50$ minutes; experiment B / D each in $\approx\!10$ minutes; the neural overlay (3 conditions $\times$ 3 seeds $\times$ 8 sample sizes) in $\approx\!9$ minutes with cached pretraining (one-time pretrain per seed: $\approx\!3.5$ minutes). All scripts seed \texttt{random}, \texttt{numpy}, and \texttt{torch} so re-runs reproduce CSVs to $10^{-8}$. Earlier development used additional CPU compute on prototype variants (different operator pools, prefix lengths $M$, and exploratory hyperparameter ranges) that did not enter the final paper; we estimate this preliminary compute at no more than a small constant multiple of the reported runtimes.

\begin{figure}[h]
    \centering
    \includegraphics[width=\linewidth]{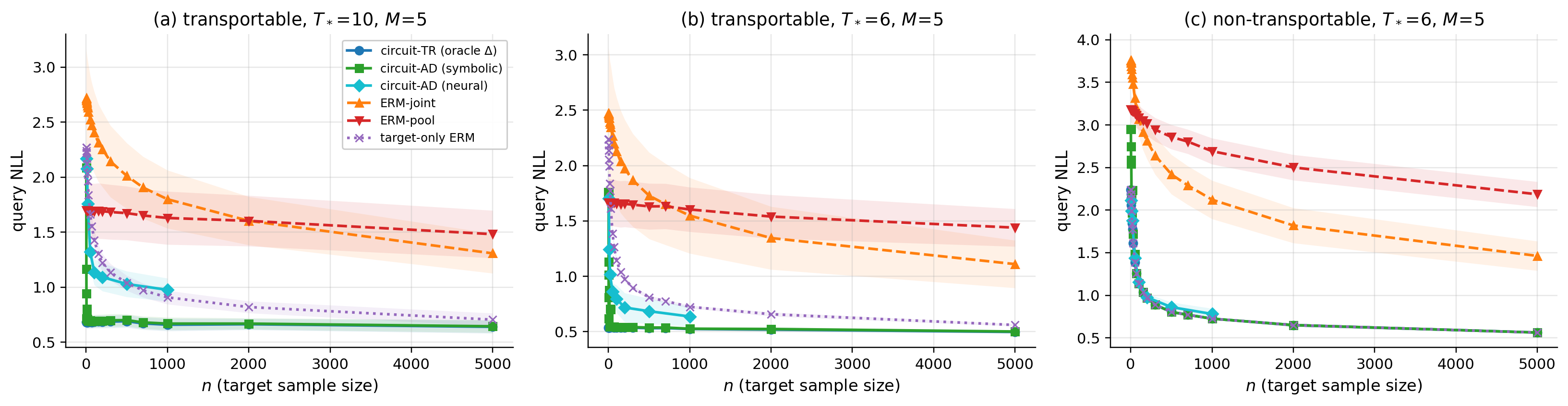}
    \caption{Full target-sample sweep ($n\!\leq\!5000$) for the three conditions of \Cref{fig:adapt-T=10}: (a) fast adaptation with process supervision, $T_*\!=\!10$; (b) fast adaptation without process supervision, $T_*\!=\!6$; (c) slow adaptation, $T_*\!=\!6$ with the last target operator outside the source pool. Mean $\pm$ 1 SE over $10$ seeds for the symbolic curves and $3$ seeds for the gradient-based relaxation; $|\cV|\!=\!10$, $N\!=\!5\!\times\!10^4$, $M\!=\!5$.}
    \label{fig:adapt-full}
\end{figure}

\paragraph{Full $n$-sweep for the three conditions.} The main text's \Cref{fig:adapt-T=10} shows $n\!\leq\!50$ for the three conditions of experiment A2; \Cref{fig:adapt-full} extends the sweep to $n\!=\!5000$. The orderings established at small $n$ persist throughout: circuit-TR/AD remain at oracle level in the two fast-adaptation panels (a) and (b) for all $n$; in the slow-adaptation panel (c) the curves of oracle circuit-TR, target-only ERM, and circuit-AD continue to coincide while the pooled-ERM baselines stay above by $1$--$2$ nats.

\begin{figure}[h]
    \centering
    \begin{subfigure}[b]{0.42\linewidth}
        \centering
        \includegraphics[width=\linewidth]{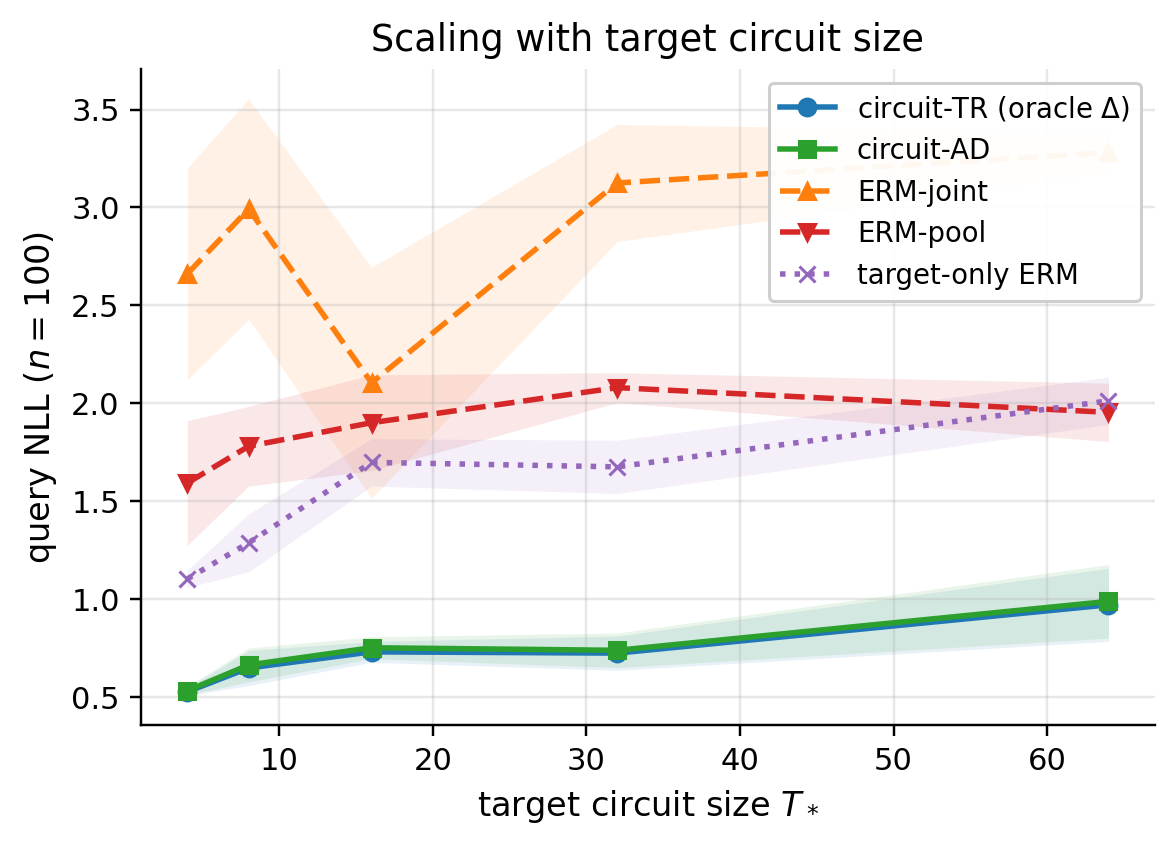}
        \caption{Scaling with target circuit size $T_*$.}
        \label{fig:T-sweep}
    \end{subfigure}\hspace{0.03\linewidth}
    \begin{subfigure}[b]{0.42\linewidth}
        \centering
        \includegraphics[width=\linewidth]{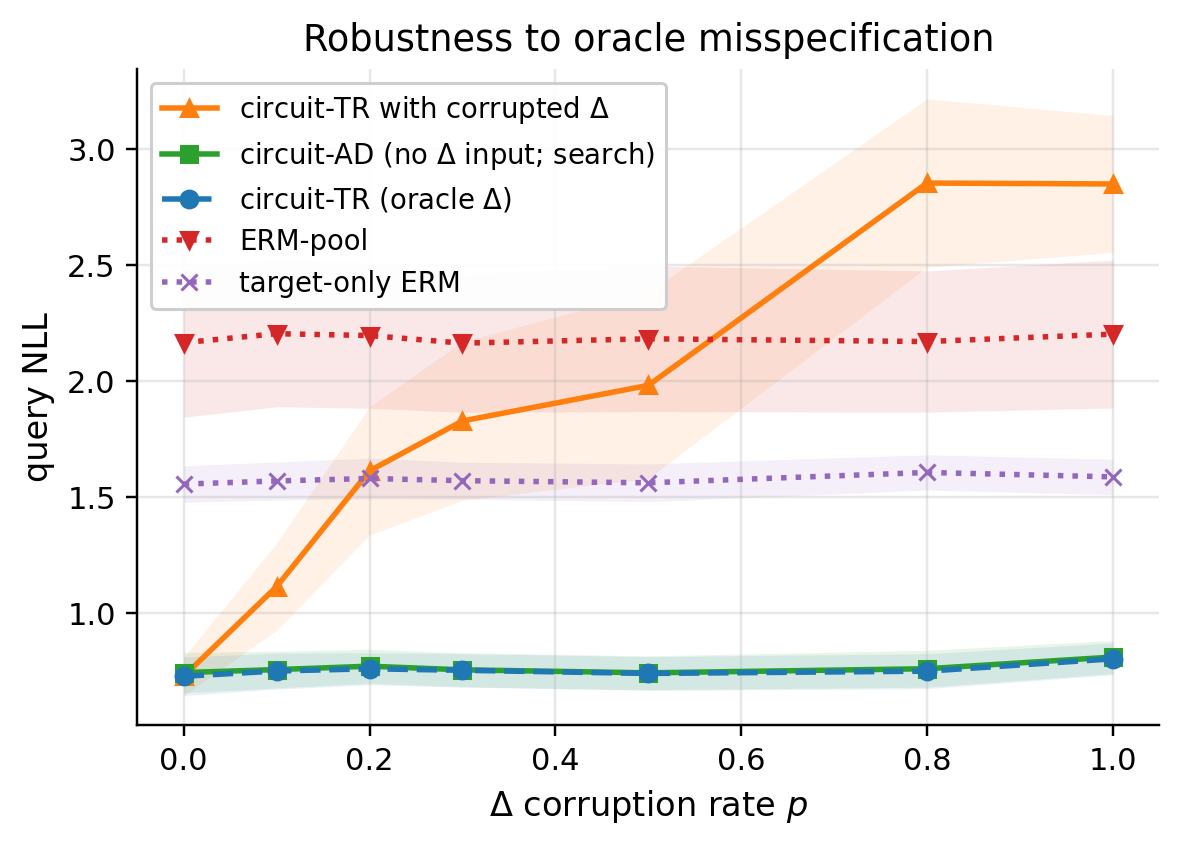}
        \caption{Robustness to corrupted oracle $\Delta$.}
        \label{fig:misspec}
    \end{subfigure}
    \caption{Stress tests. (a)~Scaling with $T_*$ at $n\!=\!100$, $|\cV|\!=\!10$, $N\!=\!5\!\times\!10^4$: target-only ERM and pooled baselines stay $0.6$--$1.0$ nats above circuit-TR/AD across $T_* \in \{4,8,16,32,64\}$. (b)~Robustness to a corrupted prior $\Delta$ at $n\!=\!100$. \emph{Circuit-TR with corrupted~$\Delta$} (orange) treats $\Delta$ as a fixed input and degrades smoothly with the corruption rate $p$. \emph{Circuit-AD} (green) takes no $\Delta$ as input and runs its internal search; its curve is essentially flat in $p$ and tracks oracle circuit-TR (blue) within the validation noise, demonstrating that the agnostic search is exactly the robustness mechanism for an unreliable would-be oracle. Bands are mean $\pm$ 1 SE over $5$ seeds.}
    \label{fig:scaling-misspec}
\end{figure}

\paragraph{Stress tests of circuit-AD.} \Cref{fig:scaling-misspec} reports two complementary stress tests referenced from the main text. Panel (a) sweeps target circuit size $T_*$ at fixed $n=100$, $|\cV|=10$, $N=5\!\times\!10^4$: circuit-TR and circuit-AD remain $0.6$--$1.0$ nats below target-only ERM and pooled baselines across the entire range $T_* \in \{4,8,16,32,64\}$. The small target-data regime makes the per-target-position effective-sample argument concrete: with $n\!=\!100$ samples and $|\cV|^{c+1}\!=\!1000$ cells per conditional, target-only ERM is sample-starved while circuit-TR/AD pool source positions to recover near-oracle accuracy. Panel (b) quantifies how unreliable a would-be oracle $\Delta$ would have to be before it stops being useful. We corrupt $\Delta_{\true}$ entrywise at rate $p$ and feed the corrupted $\Delta$ to circuit-TR as a \emph{fixed} prior. The "trust-the-prior" curve degrades smoothly from oracle level at $p\!=\!0$ ($\approx\!0.74$ nats) to worse than ERM-pool by $p\!=\!1$ ($\approx\!2.85$ nats). In contrast, circuit-AD takes \emph{no} $\Delta$ as input: its internal search over candidate clusterings (\Cref{alg:agnostic-sequence}) produces a curve that is essentially flat in $p$ and tracks oracle circuit-TR closely (small residual gap because at $n\!=\!100$ the held-out validation step has limited statistical power). This is exactly the design point of circuit-AD --- the search is the robustness mechanism for an unreliable prior.

\begin{figure}[h]
    \centering
    \includegraphics[width=0.55\linewidth]{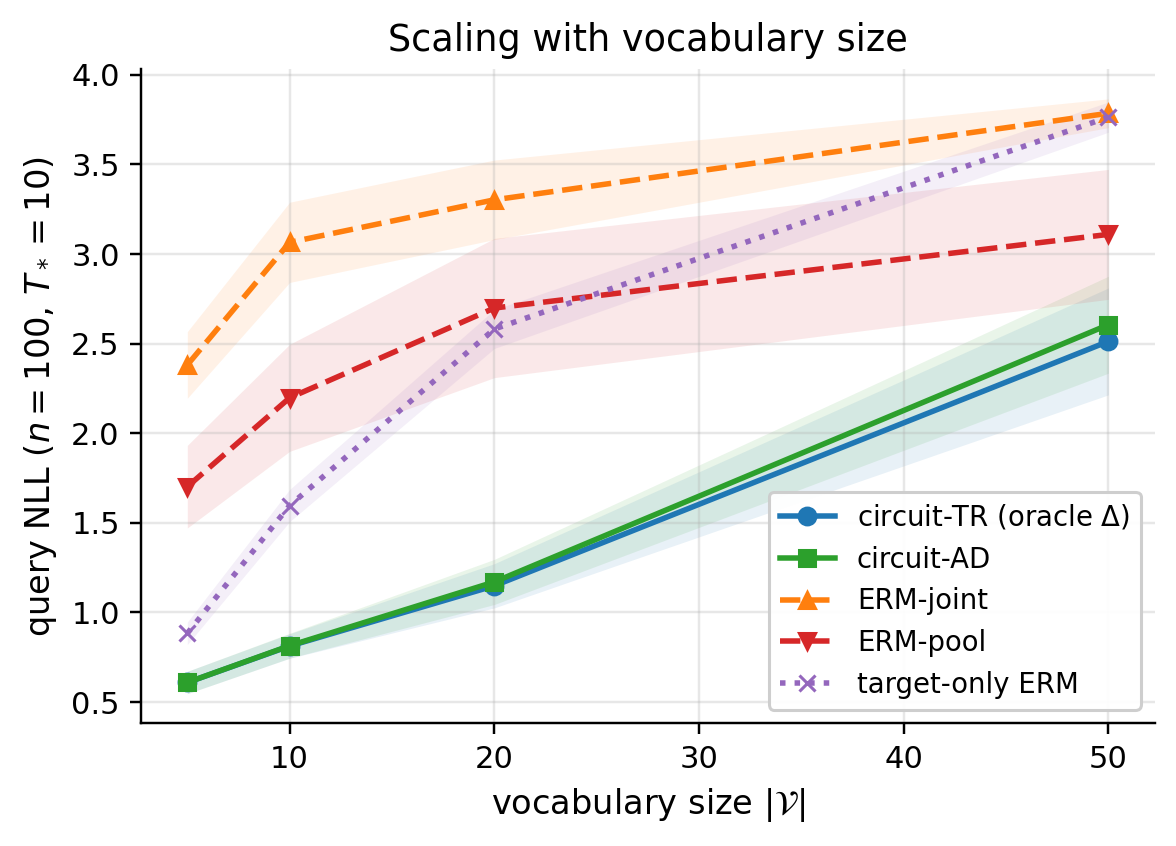}
    \caption{Vocabulary scaling at fixed $T_*\!=\!10$, $n\!=\!100$, $N\!=\!5\!\times\!10^4$. All methods worsen with $|\cV|$ as conditional tables become sparser and per-position Bayes risk increases, but circuit-TR/AD's lead over both target-only and pooled-ERM baselines grows. Error bars: 1 SE over 5 seeds.}
    \label{fig:V-sweep}
\end{figure}

\paragraph{Vocabulary scaling.} \Cref{fig:V-sweep} sweeps $|\cV| \in \{5,10,20,50\}$ at fixed $T_*\!=\!10$, $n\!=\!100$, $N\!=\!5\!\times\!10^4$. All methods worsen as $|\cV|$ grows because (i) the Bayes-optimal NLL itself grows with $|\cV|$ (per-position conditional entropy of a noisy operator is a monotone function of $|\cV|$ at fixed noise rate), and (ii) per-position conditional tables become sparser ($N/|\cV|^{c+1}\!\approx\!0.4$ samples per cell at $|\cV|\!=\!50$ with $c\!=\!2$ parents). At $n\!=\!100$, target-only ERM is severely sample-limited and circuit-TR/AD's lead grows from $0.27$ nats at $|\cV|\!=\!5$ to $\sim\!1.4$ nats at intermediate $|\cV|$. The agnostic overhead --- the gap between circuit-AD and oracle circuit-TR --- remains a small fraction of total NLL (under $0.10$ nats at $|\cV|\!=\!50$), consistent with the theory's prediction that $R_{P^\star}(\mu_\AD) - R_{P^\star}(\mu_\TR) = \cO(\sqrt{T_*^3 \log T_*/n})$ has no explicit $|\cV|$ term (\Cref{thm:agnostic-sequential}).

\subsection{Held-out GCD composition (\Cref{fig:gcd-task})} \label{sec:reproducibility-gcd}

\paragraph{Construction.} Unlike the random-SCM benchmark used in experiments A--D, the GCD experiment fixes both SCMs by hand so that the target's data-generating process is a real algorithm. The target SCM lays out the Euclidean recursion with state $(a,b)$ initialised at $V^\star_0,V^\star_1\sim\mathrm{unif}(\cV)$ and, for $k=1,\dots,5$ iterations,
\[
  V^\star_{3k-1}=\max(a_k,b_k),\quad V^\star_{3k}=\min(a_k,b_k),\quad V^\star_{3k+1}=V^\star_{3k-1}\bmod V^\star_{3k}, \qquad (a_{k+1},b_{k+1})=(V^\star_{3k},V^\star_{3k+1}),
\]
followed by a terminal $V^\star_{T_*-1}=\max(V^\star_{3K},V^\star_{3K+1})$ where $K=5$. We adopt the standard $\gcd$ convention $a\bmod 0=a$ so that once the Euclidean recursion converges (rem $=0$), the gcd is preserved at the $\min$ slot of subsequent iterations and is read off by the terminal $\max$. With $|\cV|=10$, $K=5$ iterations cover every $(V_0,V_1)\in\cV^2$ (worst-case Fibonacci-adjacent depth is $4$). Each step retains the standard $p_\mathrm{noise}=0.1$ from experiments A--D (so positivity holds), giving an effectively $\gtrsim\!90\%$ exact-gcd target distribution at the terminal position. The source SCM is a single fixed configuration of $T\!=\!15$ positions ($V^{\src}_0,V^{\src}_1,V^{\src}_2$ uniform inputs; the remaining $12$ positions instantiate the operators $\mathrm{add},\mathrm{subtract},\mathrm{mult},\min,\max,\bmod,\mathrm{copy},\mathrm{add\text{-}one},\mathrm{minus\text{-}one}$ on diverse parent pairs --- with $\min,\max,\bmod$ each appearing at two positions on different parent indices). $\Delta_{\true}$ has $26$ shared-mechanism cells across the $11\!\times\!15$ position grid; every target position with $i\!\geq\!2$ has exactly two source matches.

\paragraph{Query and metric.} The observed prefix is $M\!=\!2$ ($V^\star_0,V^\star_1$), the query is $P^\star(V^\star_{T_*-1}\mid V^\star_0,V^\star_1)$, and the held-out evaluation set has $1000$ rows. We report query NLL on this set, matching the metric used in all other experiments in the paper. \texttt{run\_neural\_gcd.py} also records an exact-gcd argmax accuracy in the output CSV for sanity-checking; the symbolic ordering of methods coincides with the NLL ordering on that metric (with target-only ERM at $\approx\!74\%$ versus circuit-AD at $\approx\!99\%$ by $n\!=\!1000$), but we omit it from the figure for visual consistency with \Cref{fig:adapt-T=10,fig:adapt-full}.

\paragraph{Sweep, seeds, and runtime.} The symbolic sweep covers $n\in\{1,2,3,5,10,20,50,100,200,500,1000,2000,5000\}$ over $5$ seeds; the gradient-based relaxation covers $n\in\{1,2,3,5,10,20,50,100,200,500,1000\}$ over $3$ seeds. \Cref{fig:gcd-task} plots only $n\in[5,1000]$: the lower bound avoids the very-small-$n$ regime where circuit-AD's held-out selection step uses $n_{\mathrm{select}}\!\leq\!1$ rows and is therefore dominated by sampling noise, and the upper bound matches the per-condition panels in \Cref{fig:adapt-T=10}. The symbolic ordering at $n\in\{2000,5000\}$ is unchanged (circuit-TR/AD remain at oracle level, target-only ERM continues to improve toward $\approx\!2.01$ nats, ERM-joint and ERM-pool plateau as in the plot). Source size $N=5\!\times\!10^4$, identical to experiments A--D. Within each $n$-cell, target data is split $70/30$ between candidate fitting and held-out selection. Symbolic methods reuse the same Laplace-smoothed-histogram estimator and $200$-rollout Monte-Carlo chain-rule evaluator. The gradient-based relaxation reuses the architecture and hyperparameters of \Cref{sec:pre-training} verbatim (hidden width $r=64$, cluster size $d=12$, AdamW lr $10^{-3}$ for pretraining and $5\!\times\!10^{-4}$ for fine-tuning); the only change from experiment A2 is that pretraining is performed once on the arithmetic-primitives source ($T\!=\!15$) instead of the random source ($T\!=\!10$), cached under \texttt{checkpoints/neural\_pretrain\_arith\_T15\_V10\_d12\_r64\_seed\{0,1,2\}.pt}. The end-to-end runtime on a single CPU core is $\approx\!3$ minutes for the symbolic sweep and $\approx\!4$ minutes for the neural overlay (cached pretrains reused across all $n$ values).

\section{Detailed model architecture}\label{app:multiple-parents-sequence}

This section provides a comprehensive walkthrough of our model architecture, focusing on how multiple parents are identified and utilized for next-token prediction in a domain-adaptive manner.

\subsubsection*{Overall task and core design principle}

The objective is sequence modeling, specifically to predict the next token in a sequence of discrete symbols (digits $0$--$9$ in our experiments). The core idea is that the generation of a token at position $i$ depends on:
\begin{enumerate}
    \item A selected \textbf{causal function} (e.g., add, subtract, multiply)
    \item One or more \textbf{parent tokens} from earlier positions in the sequence (i.e., positions $< i$)
\end{enumerate}

\subsubsection*{Input representation and positional encoding}

\textbf{Input sequences}: Sequences of integer token IDs from $\cV = \{0, 1, \ldots, 9\}$.

\textbf{Positional encoding} (\texttt{PositionalEncoding} class): Each token at position $i$ in domain $d \in \{\src, *\}$ is mapped to a dense vector representation using:
\begin{itemize}
    \item Standard sinusoidal positional encodings for the position index ($0$ to $T-1$ on source, $0$ to $T_*-1$ on target) and a domain indicator (source vs.\ target)
    \item These two embeddings (each $r/2$ dimensional) are concatenated to form the initial $r$-dimensional embedding
    \item Input: $(B, T)$ for positions and domain; Output: $(B, T, d)$ embeddings, where $B$ is the batch size
\end{itemize}

This design lets the model condition on both absolute position and the source/target domain, enabling per-domain parent selection.

\subsubsection*{Universal operator indicator}

The \texttt{UniversalOperatorIndicator} class determines, for each token position, the probability distribution over a fixed set of operations $\cF$ (e.g., add, subtract, multiply\_two).

\textbf{Mechanism}:
\begin{enumerate}
    \item The $h$-dimensional embedding of each token is passed through a linear projection to $|\mathcal{F}|$ dimensions
    \item Layer normalization is applied: $\text{LayerNorm}(\text{Linear}(\text{embedding}))$
    \item Softmax produces a probability distribution: $\text{softmax}(\text{LayerNorm}(\text{Linear}(\text{embedding})))$
\end{enumerate}

\textbf{Universality}: This module's parameters are shared across all domains and positions. The \emph{choice} of operation is contextual based on the token's embedding, but the \emph{meaning} of each operation is universal across domains.

\textbf{Output}: $(B, T, |\mathcal{F}|)$ operator probabilities where $B$ is batch size.

\subsubsection*{Domain-specific parent selector}

The \texttt{DomainSpecificParentSelector} class implements the core mechanism for identifying influential parent tokens, corresponding to the causal structure learning described in \Cref{alg:agnostic-sequence}.

\textbf{Multi-head causal attention design}: The mechanism uses $C$ distinct attention heads (where $C$ is the maximum number of parents). For our experiments, $C = 2$, meaning the model can identify up to two distinct parents for each token.

\textbf{Domain-specificity}: The key innovation is that query ($\query$) and key ($\key$) projection matrices are domain-specific:
\begin{align}
    \text{domain\_queries} &\in \mathbb{R}^{2 \times C \times r \times r} \\
    \text{domain\_keys} &\in \mathbb{R}^{2 \times C \times r \times r}
\end{align}
where the leading $2$ indexes source vs.\ target. During forward pass, for domain $d \in \{\src, *\}$ and parent head $h$:
\begin{align}
    \query_{d,h} &= \text{Embeddings} \cdot W^{\query}_{d,h} \\
    \key_{d,h} &= \text{Embeddings} \cdot W^{\key}_{d,h}
\end{align}

\textbf{Parent selection process}: For each domain $d$, parent head $h$:
\begin{enumerate}
    \item \textbf{Attention scores}: $S_{d,h} = \frac{\query_{d,h} \key_{d,h}^T}{\sqrt{r}}$
    \item \textbf{Causal masking}: Standard causal attention masking ensures token at position $i$ only attends to positions $< i$
    \item \textbf{Sharp softmax}: $A_{d,h} = \text{softmax}(S_{d,h} / \tau)$ where $\tau = 0.1$
    \item \textbf{First position handling}: Weights for position $0$ are zeroed as it has no parents
\end{enumerate}

The temperature $\tau = 0.1$ makes the softmax significantly sharper, encouraging sparse selection of a small number of parents rather than soft averaging.

\textbf{Output}: For each domain $d \in \{\src, *\}$, a list of $C$ attention weight matrices $(B, T, T)$ representing parent selection distributions.

\subsubsection*{Feature preparation for conditional MLP}

The \texttt{\_prepare\_mlp\_features} method combines operator indicators and selected parent values to form input for the final prediction MLP.

\textbf{Input components}:
\begin{itemize}
    \item $\text{sequences\_onehot} \in \{0,1\}^{B \times T \times |\cV|}$: One-hot encoded input sequences
    \item $\text{operator\_indicators} \in [0,1]^{B \times T \times |\mathcal{F}|}$: From universal operator indicator
    \item $\text{parent\_weights}$: From domain-specific parent selector
    \item $\text{domains} \in \{\src, *\}^{B \times T}$: Source/target indicator
\end{itemize}

\textbf{Feature construction}: For each position $i$:
\begin{enumerate}
    \item Operator indicators $\text{operator\_indicators}[:,p,:]$ form the first part of the feature vector
    \item For each parent head $h \in \{0, \ldots, C-1\}$:
        \begin{equation}
            \text{WeightedParentValue}_{h,p} = A_{d,h}[:,p,:] \cdot \text{sequences\_onehot}
        \end{equation}
        where $d$ is the domain of position $i$. This produces a $(B, |\cV|)$ vector for each parent head.
    \item These $C$ vectors are concatenated after the operator indicators
\end{enumerate}

\textbf{Feature dimension}: $|\mathcal{F}| + (C \times |\cV|)$ where $|\mathcal{F}|$ is the number of operations and $|\cV| = 10$.

\subsubsection*{Conditional MLP for prediction}

The \texttt{EfficientConditionalMLP} class predicts the next token's probability distribution based on the combined features, implementing the learned conditional distributions $\hat{P}(v_i | \text{pa}_i)$ from our theoretical framework.

\textbf{Architecture}:
\begin{enumerate}
    \item \textbf{Input projection}: Linear layer from feature dimension to $r$
    \item \textbf{Hidden layers}: Stack of linear layers ($r \to r$) with ReLU activations, dropout, and residual connections
    \item \textbf{Output layer}: Linear layer from $r$ to $|\cV| = 10$
\end{enumerate}

\textbf{Output}: Logits of shape $(B, T, |\cV|)$ for next-token prediction.

\subsubsection*{Training and fine-tuning protocol}

\textbf{Pre-training (source domains)}: The entire model, including domain-specific $\query/\key$ matrices for source domains, is trained end-to-end using standard next-token prediction cross-entropy loss.

\textbf{Fine-tuning (target domain adaptation)}: When adapting to target domain $\pi^\star$:
\begin{itemize}
    \item \textbf{Frozen components}: 
        \begin{itemize}
            \item Positional encoding parameters
            \item Universal operator indicator parameters  
            \item Conditional MLP parameters
            \item Source domain $\query/\key$ matrices
        \end{itemize}
    \item \textbf{Trainable components}:
        \begin{itemize}
            \item New randomly initialized $W^\query_{*,h}, W^\key_{*,h}$ for target domain
            \item New target-specific operator indicator
        \end{itemize}
\end{itemize}

This modular design enables learning domain-specific parent selection while reusing universal causal functions, corresponding to the structure-agnostic adaptation strategy in \Cref{alg:agnostic-sequence} with computational efficiency discussed in \Cref{sec:optimization}.
\section{Related work in compositional generalization} \label{app:related-comp-gen}

The bulk of this paper sits in the causal-inference literature on transportability and domain adaptation. We briefly acknowledge two adjacent lines that share surface vocabulary with our setting; the discussion is for the reader's orientation, not a claim of contribution to those literatures.

\paragraph{Compositional generalization in seq2seq / parsing.} A substantial line studies whether neural models systematically recombine seen primitives within a single distribution: SCAN \citep{lake2018scan}, COGS \citep{kim2020cogs}, gSCAN \citep{ruis2020gscan}, and the survey by \citet{hupkes2020compositionality}. These benchmarks construct a single generating grammar and evaluate held-out compositions of seen primitives -- a domain-generalization (no target data) problem in our terminology, with \emph{primitive} typically referring to a deterministic atomic vocabulary token. Meta-learning extensions \citep{lake2019meta,conklin2021meta} either run in-context over support sets without parameter adaptation or use meta-learning as a single-distribution regularizer; neither corresponds to our few-shot transfer setting with a structurally distinct target.

\paragraph{Algorithmic composition.} A separate line studies composition of subroutines in algorithmic tasks: Neural Module Networks \citep{andreas2016neural}, the CLRS algorithmic-reasoning benchmark \citep{velivckovic2022clrs} and its multi-task transfer study \citep{xhonneux2021transfer}, length generalization in arithmetic transformers \citep{zhou2024algorithms,anil2022length}, and library-learning systems such as DreamCoder \citep{ellis2021dreamcoder}. These works share the spirit of our setting -- primitive operations reused at multiple positions in a target circuit -- but their primitives are typically deterministic and given (or symbolically synthesized), and the formal sample-complexity story is largely empirical. Our \emph{modules} are probabilistic and learned from a source distribution, with sample-complexity rates governed by the size of the target circuit composable from those modules.

\paragraph{Independent causal mechanisms and modular learning.} \citet{parascandolo2018learning} recover inverse mechanisms unsupervisedly when one canonical distribution has been transformed by $N$ independent operators; the setting differs structurally from ours (no source/target distinction, no causal graph, no discrepancy oracle) but shares the high-level idea that mechanisms are autonomous modules.

We make no claim of contribution to the comp-gen, NAR, or program-induction literatures cited above. The present work's home is causal inference and transportability theory; we point to the adjacent lines so that readers coming from them can locate the differences in problem formulation.

\section{Limitations} \label{app:limitations}
While our work presents a novel causal framework for domain adaptation in sequence learning, several limitations should be acknowledged:

\begin{enumerate}
    \item \textbf{Discrete variables}: Throughout this paper, we focus exclusively on discrete-valued variables. Extending our framework to continuous variables would require different mathematical techniques and might introduce additional challenges in estimating causal effects.

    \item \textbf{Strict positivity assumption.} Leveraging tools from statistical learning theory and causal inference in the way presented in this work relies on the positivity assumption ($P(x,y) > \epsilon$), that is a constant baseline probability of observing any sequence of tokens. In many real-world settings, there exists strict rules governing the distribution of sequences, e.g., certain tokens necessarily follow after some others, thus, this assumption may be violated in presence of determinism.
    
    \item \textbf{Computational scalability.} The symbolic structure-agnostic algorithm (\Cref{alg:agnostic-sequence}) explores all possible partitions of position-domain pairs and causal diagrams, which becomes computationally intractable as the number of variables grows. The gradient-based relaxation in \Cref{sec:optimization} mitigates this in practice, but scalability remains a concern for long sequences.
    
    \item \textbf{Presence of confounders}: We discuss confounding in \Cref{app:confounders}, and illustrate how it complicates the problem. Our main theoretical results and algorithms primarily address scenarios without unobserved confounding. Many real-world sequences may exhibit more complex confounding patterns that could limit direct application. Moreover, we assume a causal ordering of variables, which is the nature of sequential data; in much of structure learning literature finding a correct causal order is the key challenge.

    \item \textbf{Evaluation on real algorithmic traces vs.\ natural-language benchmarks.} Beyond random SCMs from an operator pool, we include in \Cref{fig:gcd-task} an experiment whose target data-generating process is a real algorithm (Euclidean's GCD), with $\cG^\star$, $\cG^{\src}$, and $\Delta$ dictated by the algorithm itself rather than fabricated --- direction (a) below. This addresses the most common ``synthetic-only'' objection by exercising the framework on a held-out arithmetic composition, where circuit-AD recovers near-oracle performance and outperforms both target-only ERM and the pooled-ERM baselines at the same target sample size. We have not extended to natural-language compositional-generalization benchmarks (SCAN, COGS, gSCAN, etc.) because for those neither $\cG^{\src},\cG^\star$ nor a meaningful $\Delta$ is available --- the ``true'' causal graph behind a sentence is itself a research question. Two principled directions for further real-world evaluation are: (a) more tasks where the causal structure is grounded by the data-generating process (other program-execution traces, controlled physical simulations, structured tabular pipelines), and (b) using a frontier LLM as a noisy proposer of $\cG^{\src},\cG^\star$ and $\Delta$, then quantifying how circuit-AD performs as a function of oracle quality. We see (b) as the most direct path from this work to language-style benchmarks but as out of scope for the present paper, which focuses on establishing the theory.

    \item \textbf{Theory-practice gap in the gradient relaxation.} The error rate guarantees (\Cref{thm:agnostic-sequential,thm:structure-informed-sequence}) are stated for the symbolic circuit-AD algorithm; the gradient-based architecture in \Cref{sec:optimization} is an empirical relaxation, and we currently provide no formal result that ties its solutions to circuit-AD's enumeration. Closing this gap is the main open theoretical problem the framework raises.
\end{enumerate}

Beyond these specific limitations, the framework is best read as identifying \emph{which} aspects of compositional generalization are theoretically transportable when the causal structure is known. Operationalizing it in the wild --- where structure must be learned, partially specified, or solicited from another model --- is a natural next research arc rather than a single follow-up experiment.

\end{document}